\documentclass[twoside,11pt]{article}

\usepackage{jmlr2e}
\usepackage{comment}

\jmlrheading{1}{2000}{1-48}{4/00}{10/00}{Marina Meil\u{a} and Michael I. Jordan}

\ShortHeadings{What Large Language Models Know and What People Think They Know}{What Large Language Models Know and What People Think They Know}
\firstpageno{1}

\begin{document}

\title{What Large Language Models Know and What People Think They Know}

\author{\name Mark Steyvers \email mark.steyvers@uci.edu \\
       \addr Department of Cognitive Sciences,
       University of California, Irvine
       \AND
       \name Heliodoro Tejeda \email htejeda@uci.edu \\
       \addr Department of Cognitive Sciences,
       University of California, Irvine
       \AND
       \name Aakriti Kumar \email aakritk@uci.edu \\
       \addr Department of Cognitive Sciences,
       University of California, Irvine
        \AND
       \name Catarina Belem \email cbelem@uci.edu \\
       \addr Department of Computer Science,
       University of California, Irvine
       \AND
       \name Sheer Karny \email skarny@uci.edu \\
       \addr Department of Cognitive Sciences,
       University of California, Irvine
       \AND
       \name Xinyue Hu \email xhu26@uci.edu \\
       \addr Department of Cognitive Sciences,
       University of California, Irvine
       \AND
       \name Lukas Mayer \email lwmayer@uci.edu \\
       \addr Department of Cognitive Sciences,
       University of California, Irvine
       \AND
       \name Padhraic Smyth \email smyth@ics.uci.edu \\
       \addr Department of Computer Science,
       University of California, Irvine
       }

%\editor{Leslie Pack Kaelbling}

\maketitle

\begin{abstract}%   <- trailing '%' for backward compatibility of .sty file
As artificial intelligence systems, particularly large language models (LLMs), become increasingly integrated into decision-making processes, the ability to trust their outputs is crucial. To earn human trust, LLMs must be well calibrated such that they can accurately assess and communicate the likelihood of their predictions being correct. Whereas recent work has focused on LLMs’ internal confidence, less is understood about how effectively they convey uncertainty to users. Here we explore the calibration gap, which refers to the difference between human confidence in LLM-generated answers and the models’ actual confidence, and the discrimination gap, which reflects how well humans and models can distinguish between correct and incorrect answers. Our experiments with multiple-choice and short-answer questions reveal that users tend to overestimate the accuracy of LLM responses when provided with default explanations. Moreover, longer explanations increased user confidence, even when the extra length did not improve answer accuracy. By adjusting LLM explanations to better reflect the models’ internal confidence, both the calibration gap and the discrimination gap narrowed, significantly improving user perception of LLM accuracy. These findings underscore the importance of accurate uncertainty communication and highlight the effect of explanation length in influencing user trust in artificial-intelligence-assisted decision-making environments.
\end{abstract}


\section{Introduction}
Uncertainty communication plays a critical role in decision-making and policy development. Uncertainties are often expressed verbally to help stakeholders understand risks and make informed choices across a wide range of domains, including climate policy, law, medicine, and intelligence forecasting. Psychology research has investigated perceptions of verbally expressed uncertainty (e.g., phrases such as ``very unlikely'', or ``almost certain'') in these domains \cite{budescu2014ipcc, ho2015communicating, karelitz2002toward, wallsten2008intel,o1989words}. Despite their lack of precision in communicating probabilities, verbal probability phrases provide a simple and effective way to communicate uncertainty in natural language contexts. The emergence of large language models (LLMs) introduces new complexities in the area of uncertainty communication. These models are increasingly integrated into  
areas such as public health \cite{ali2023using}, coding \cite{zambrano2023ncoder}, and education \cite{whalen2023chatgpt}. However, the question of how effectively LLMs communicate uncertainty is unexplored. As the primary mode of communication with LLMs is through natural language, it is critical to understand if LLMs are able to accurately convey through verbal means what they know or do not know.    

Recent research raises doubts about the reliability of the information that LLMs generate. One notable issue is the possibility of generating responses that, while convincing, may be inaccurate or nonsensical \cite{jo2023promise,huang2023survey}. The unreliability of LLMs has led developers of LLMs to caution against uncritical acceptance of model outputs \cite{openai2022chatgpt}, suggesting that it is not always clear when the models are or are not confident in the knowledge communicated to the user.

At the same time, recent research has also indicated that LLMs have the ability, to a certain degree, to accurately discern their own knowledge boundaries. LLMs in particular can exhibit a reasonable level of calibration for multiple-choice questions such that the probability the model assigns to a selected answer tracks with the probability that this answer is correct \cite{openai2023gpt4,kadavath2022language,srivastava2023beyond}. In addition, recent studies show that LLMs can distinguish between answerable and unanswerable questions \cite{yin2023large, kadavath2022language} and the internal state of an LLM  can distinguish between truthful statements and lies \cite{azaria2023internal} and truthful statements and confabulations \cite{farquhar2024detecting}. These findings suggest that LLMs may possess an internal mechanism that is reflective of self-knowledge.

In the specific context of question-answering, an LLM's ``model confidence'' is typically equated to the probability assigned by the LLM to the selected answer relative to other possible answers (e.g., \cite{jiang2021can,hendryckstest2021}). However, from the perspective of a human interacting with the LLM, this internal model confidence is not usually displayed to human users as part of LLM output.  Instead, in current practice, humans rely solely on the language produced by the LLM in order to assess LLM confidence. To contrast with model confidence, in this paper we use the term ``human confidence''  to refer to a human's assessment (expressed as a probability) of how likely it is that the LLM's answer is correct based only on the language produced by the LLM without any knowledge of the LLM's internal model confidence.

Surprisingly, studies focused on investigating human confidence in LLM outputs are lacking. In this paper, we take a step in addressing this issue and investigate what we term ``the calibration gap'', namely the difference in the reliability of LLM model confidence and human confidence. In effect, the calibration gap represents the gap between an LLM's own internal confidence of what it knows and human perception of this confidence. In addition, we investigate ``the discrimination gap'', which relates to the difference in the ability to discriminate between likely correct and incorrect answers. Any discrimination gap shows that whatever internal LLM representation is used to tell the difference between likely correct and incorrect answers is not conveyed effectively to humans. We address two specific research questions in this context. First, how large are the calibration and discrimination gaps? i.e., is there a significant gap between LLM model confidence and human confidence in terms of how each assesses the true accuracy of the LLM? Second, can the calibration and discrimination gaps be reduced? Can the quality of human confidence in an LLM be improved by adapting the textual output of the LLM to internal model confidence? These questions have important implications for the design of reliable LLM assistants. By aligning the LLM's internal confidence with human perception of this confidence, we can bridge the gap between what LLMs know and what people think they know, which is crucial for the development of effective and trustworthy assistants 

Our contributions in this context are twofold. First, we present a set of experimental studies and dataset that directly captures human assessment of LLM confidence in a question-answering context, providing insight into human perceptions of LLM textual responses. Second, we test and suggest ways of generating LLM responses that improve the calibration quality of human confidence relative to the LLM assistant's model confidence and the LLM's true accuracy.

\subsection{Large Language Models}
We use three publicly available LLMs in our studies: GPT-3.5 \cite{GPT-3.5INSTRUCT}, PaLM2 \cite{anil2023palm}, and GPT-4o. We apply the GPT-3.5 and PaLM2 models to a subset of multiple-choice questions from the Massive Multitask Language Understanding (MMLU) dataset, a comprehensive dataset that contains multiple-choice questions from various knowledge domains, such as STEM, humanities, social sciences, and more \cite{hendryckstest2021}.  We apply the GPT-4o model to a subset of short-answer questions from the Trivia QA data set \cite{joshi2017triviaqa}. For each multiple-choice and short-answer question, we assess model confidence by computing the token likelihoods (see Methods for details). This method for reading out model confidence allows for a direct computation of the relative probabilities of different possible answers in multiple-choice questions \cite{jiang2021can, kadavath2022language, xiao2022uncertainty, hendryckstest2021,openai2023gpt4} and the probability that the answer to an open-ended question is correct \cite{kadavath2022language, farquhar2024detecting}. We investigate the relationship between model confidence and accuracy to determine whether the LLM is reasonably well-calibrated, independent of the LLM's ability to elicit well-calibrated confidence from humans who use the LLM. 

\section{Methodology}

\begin{figure}[h!]
\includegraphics[width=\textwidth]{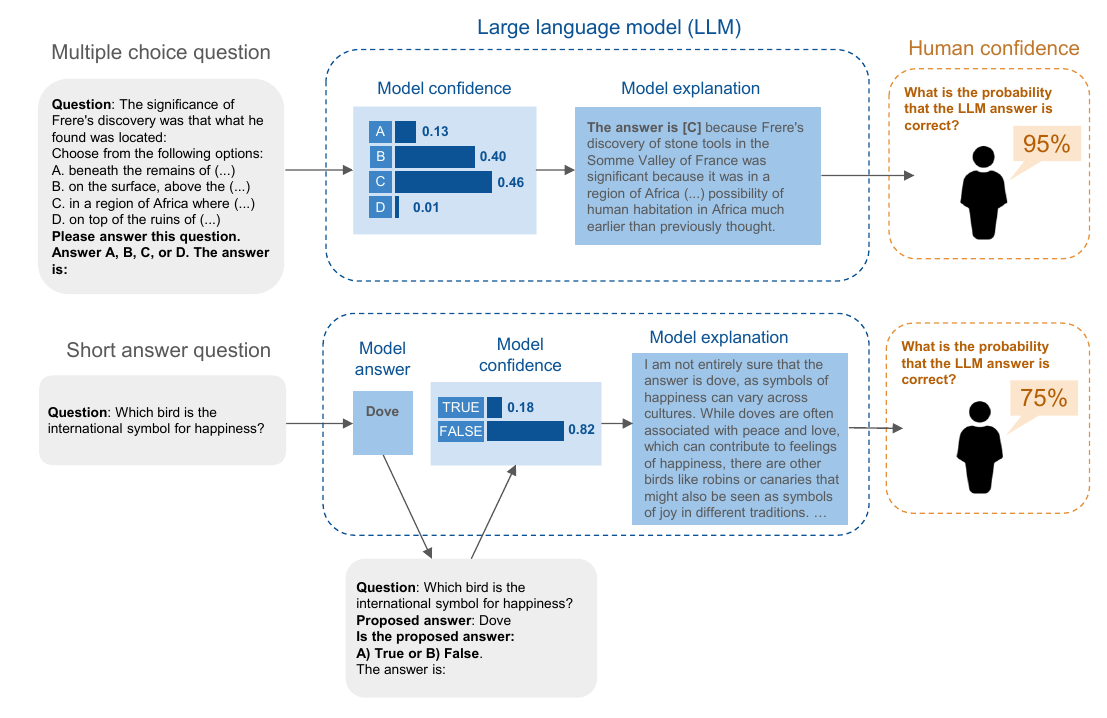}
\caption{Overview of the evaluation methodology for assessing the calibration gap between model confidence and human confidence in the model. The multiple choice questions (top), the approach works as follows: (1) prompt the LLM with a question to obtain the model's internal confidence for each answer choice; (2) select the most likely answer and prompt the model a second time to generate an explanation for the given answer; (3) obtain the human confidence by showing users the question and the LLM's explanation and asking users to indicate the probability that the model is correct. In this toy example the model confidence for the multiple choice question is 0.46 for answer C, whereas the human confidence is 0.95. For short-answer questions, the approach is similar except that internal model confidence is obtained by an additional step where the LLM is prompted to evaluate whether the previously provided answer to the question is true or false \cite{kadavath2022language}. In the short-answer question example, the LLM model explanation was modified with uncertainty language to convey the low model confidence (0.18). For the two toy examples, the correct answers are ``A'' and ``blue bird''.}
\label{fig:experimentsetup-overview}
\end{figure}

We designed behavioral experiments to evaluate human perceptions of LLM confidence. In these experiments, participants estimate the probability that the LLM's answer to a multiple-choice or short-answer question is correct based on the explanation that the LLM provided (see Figure \ref{fig:experimentsetup-overview}). Participants are not provided any direct access to the LLM's numerical model confidence, allowing us to make inferences about participants' perceptions of the confidence of the LLM based on model explanations alone. In addition, for the multiple-choice questions part of the experiment only, with the assistance of the LLM, participants provided answers to the questions. Previous research has demonstrated that the MMLU multiple-choice questions are difficult for participants who lack domain expertise, resulting in near-chance accuracy \cite{hendryckstest2021}. We anticipate that these questions will be difficult to answer without the assistance of the LLM because the majority of the participants in our experiments lack domain expertise, and their perception of the explanation's content will influence their evaluation more than their own knowledge.

We conducted two experiments each involving the three types of LLMs and two types of questions (see Table \ref{tab:overviewexperiments} for an overview). Experiment 1 assesses human perceptions of LLM accuracy using the LLM's default explanations for either multiple-choice or short-answer questions. The results from this experiment allow us to address the first research question regarding the size of the calibration and discrimination gap between model and human confidence. Experiment 2 manipulates the prompts to produce three levels of uncertainty language (low, medium, and high confidence) and three levels of explanation length, resulting in nine different types of explanations presented to participants. The prompts are designed to include uncertainty language corresponding to model confidence at the start of the explanation. Table \ref{tab:explanationexamples} illustrates explanations from a particular multiple-choice question used in the experiments (see Supplementary Table 2 for the full model explanations). The results from this experiment serve two purposes. First, we establish that human confidence varies with the uncertainty language and the length of the explanation. Next, we use the results from Experiment 2 to answer the second research question, which is to understand how the calibration and discrimination gap can be reduced by aligning the uncertainty language with model confidence—showing a low/medium/high confidence explanation when the model has low/medium/high confidence. The Supplementary Information (``Experiment 3'') reports the results from an additional experiment with a different prompting approach that alters the default explanations from Experiment 1. We use the two metrics to assess the relationship between human and model confidence and model accuracy. See Section \ref{sec:metrics} Metrics for details.

\subsection{Question data sets}
\label{sec:question-dataset}
\paragraph{MMLU dataset for multiple choice questions.} The MMLU dataset is a comprehensive multitask dataset that contains multiple-choice questions from various knowledge domains, such as STEM, humanities, social sciences, and more \cite{hendryckstest2021}. In total, there are 14042 test set questions from 57 categories curated by undergraduate and graduate students from freely available online resources such as the GRE and USMLE. These questions range in difficulty from high-school to the professional level. The MMLU dataset is widely employed to measure a text model's multitask accuracy, as it challenges models on their real-world text understanding beyond mere linguistic comprehension, thus making it a robust benchmark for model evaluation \cite{hendryckstest2021, hoffmann2022diego,rae2021scaling}. For this research, we sampled a subset of 350 questions from a range of model confidence levels in 10 select categories from the full dataset to comprehensively evaluate people's assessment of LLM model confidence.

\paragraph{Trivia QA dataset for short answer questions.} Trivia QA is a data set of trivia questions that can be answered in short answers \cite{joshi2017triviaqa}. Similar to methodology by \cite{farquhar2024detecting}, contextual information was excluded to make the question answering more challenging for LLMs and more suitable for our behavioral experiments. For this research, we assessed model confidence for 5000 questions from the original 650K dataset before selecting a final sample of 336 questions from a range of model confidence levels. The final set of questions was categorized into 7 different topics (culture \& society, entertainment, geography, history, politics, science \& technology and sports). 

\subsection{Assessing model confidence and creating question subsets}
Several approaches have been developed to elicit confidence in LLMs and to assess the degree to which the elicited confidence scores are calibrated (see \cite{geng2023survey} for an overview). In this research, we use an approach commonly used to access internal model information based on token likelihoods, allowing for direct computation of relative probabilities of different possible answers in multiple-choice questions \cite{jiang2021can, kadavath2022language, xiao2022uncertainty, hendryckstest2021,openai2023gpt4}. In addition, the token-likelihood approach can be extended to short-answer questions such that the token-likelihood reflects the model confidence that the LLM answer is correct \cite{kadavath2022language}.   

Methods that do not require access to internal model representations have used prompting strategies designed to elicit verbal expressions of uncertainty \cite{xiong2023can, zhou2023navigating}. Confidence is expressed in natural language as numeric strings (e.g., ``80\%'') \cite{lin2022teaching,xiong2023can} or more qualitative expressions of confidence (e.g., ``I am not confident the answer is X''). Prompts can be designed to emphasize step-by-step reasoning about the correctness of individual steps and clarify the space of possible answers lead to better calibration than simple prompts that simply ask for a confidence rating \cite{xiong2023can}. For short-form question answering, prompting strategies can lead to calibrated confidence levels \cite{tian2023just}. However, prompting approaches have been found to be less accurate compared to methods that read out model confidence \cite{xiong2023can}. 

\paragraph{Multiple choice questions.} For the multiple choice questions, we followed the procedures based on reading out the internal token likelihoods as described in the GPT-4 Technical Report \cite{openai2023gpt4}. We used a zero-shot prompting approach, in which the model was only prompted with the target question and its associated answer options (Extended Data Figure 1). We first assessed the LLM model confidence of GPT-3.5 and PaLM2 language models to 14042 MMLU multiple-choice questions. This allowed us to then select questions with (somewhat) evenly distributed confidence levels. We read out the log-probabilities for the top 5 tokens completed by the model using the APIs for the GPT3.5 (\texttt{gpt-3.5-turbo-instruct}) and the PaLM2 (\texttt{text-bison@002}) models. The temperature parameter was set to 0. The answer was considered complete if the tokens included the single letters A, B, C, and D. The log scores were then converted and normalized to probabilities across the four answer options (so that the sum of the scores equaled one). In this research, internal uncertainties, referred to in this paper as ``model confidence'', were represented by these probabilities in all experiments, a common technique in calibration assessment with LLMs \cite{jiang2021can, kadavath2022language, xiao2022uncertainty, hendryckstest2021,openai2023gpt4}.

%\footnote{The GPT3.5 model was accessed on Aug-Oct 2023 and PaLM2 model on January 2024.}

Based on the model confidence levels of each LLM for all MMLU questions, we created a subset separately for each LLM. In total, 35 questions were sampled for each of 10 topics, for a total of 350 questions. For each topic, the 35 questions were sampled to approximately create a uniform distribution over model confidence using the confidence bins: 0.2-0.4, 0.4-0.6, 0.6-0.8, and 0.8-1.0. However, due to the small number of questions that lead to model confidence in the lowest confidence bin, fewer questions were sampled for the 0.2-0.4 confidence range. Supplementary Figure 1 shows the distribution over model confidence levels for the entire MMLU dataset as well as the question subset sampled for our study. Model accuracy across the 350 questions is 55\% and 50\% for GPT-3.5 and PaLM2, respectively.  

\paragraph{Short-answer questions.} For the short-answer questions, we used a procedure based on the ``pTrue'' method \cite{kadavath2022language} to assess internal model confidence. All experiments with short-answer questions were performed with the API for the GPT-4o model (\texttt{gpt-4o-mini}) with the temperature parameter set to 0.7 (similar to \cite{kadavath2022language} and \cite{farquhar2024detecting}). The model was first prompted to generate the answer to each of the 5000 trivia questions in the sample. To ensure that the model response was restricted to short answers, we used a 10-shot prompting approach where the prompt contained the target question preceded by a random sample of 10 trivia question with the reference answers. Median answer length was 2 words. 

%\footnote{The GPT-4o model was accessed on Aug-Sept 2024.}

To assess model confidence for short-answer questions, as shown in Figure 1 (bottom panel), we prompted the model with the question, the proposed answer, and asked it to determine whether the proposed answer is true or false (see Extended Data Figure 1 for an example of the exact prompt). The log scores for the true and false answer options were then converted and normalized to probabilities across the two answer options. Model confidence in our experiments corresponds to the probability for the true answer option. 

For the behavioral experiments, we created a subset of 336 questions to ensure a uniform distribution across four confidence bins: 0-0.25, 0.25-0.50, 0.50-0.75, and 0.75-1.0. Supplementary Figures 1 and 2 show the distribution of model confidence levels for the 5000 sample and the 336 subset used in our behavioral experiments. Model accuracy across the 336 questions is 63\%.

We used both automatic and human scoring methods to assess model accuracy. For the 5000 question sample, we prompted an LLM (GPT-4o) to determine whether the reference answer from the Trivia QA had the same meaning as the LLM answer within the context of the question. For the 336 question sample, we additionally applied human scoring. For 97\% of questions, automatic and human scoring agreed. Model accuracy for the 336 question subset was based on human evaluation. 

\subsection{Behavioral Experiments}
\label{sec:behavioralexperiments}
This section describes the methodology we used for our behavioral experiments. Experiment 1 presented default explanations from LLMs to participants, whereas Experiment 2 presented explanations that were altered by different types of uncertainty language and overall length (see Table \ref{tab:overviewexperiments} for an overview of all experiments). Within each experiment, across different groups of participants, we varied the type of question as well as the type of LLM. Experiments 1a and 2a used explanations from GPT-3.5 for the MMLU multiple questions. Experiments 1b and 2b used explanations from PaLM2 for the MMLU multiple questions. Finally, Experiments 1c and 2c used explanations from GPT-4o for the Trivia QA short-answer questions.  The Supplementary Information (``Experiment 3'') describes the results from an additional Experiment 3, which was conducted to verify that our results generalize to different ways to vary the type of uncertainty language in the explanations.

\begin{table}[h]
  \caption{Overview of experiments.}
  \label{tab:freq}
  \centering
  \begin{tabular}{lllll}
    \toprule
   Experiment & Question Type & LLM & Explanation Type & Number of Participants\\
    \midrule
    1a & Multiple Choice & GPT-3.5 & Default explanations & 41\\
    1b & Multiple Choice & PaLM2 & Default explanations & 39\\
    1c & Short Answer & GPT-4o & Default explanations & 42\\
      \addlinespace
    2a & Multiple Choice & GPT-3.5 & Modified explanations & 60\\
    2b & Multiple Choice & PaLM2 & Modified explanations & 60\\
    2c & Short Answer & GPT-4o & Modified explanations & 59\\
  \bottomrule
\end{tabular}
\label{tab:overviewexperiments}
\end{table}

\subsubsection{Participants}
A total of 301 participants completed the study across Experiments 1 and 2 (Table \ref{tab:overviewexperiments} shows the breakdown by experiment). Participants were native English speakers residing in the United States, recruited through Prolific (www.prolific.com). Demographic data was obtained for 284 participants. There were 146 female and 138 Male participants. The median age was 34 (age range from 18 to 79). The University of California, Irvine Institutional Review Board (IRB) approved the experimental protocol. Participants who completed Experiments 1a, 1b, 2a, or 2b were paid \$8 USD for their participation. Participants in Experiments 1c and 2c required less time to complete the study and were paid \$5. The payments across experiments corresponded to a rate of approximately \$12/hr. Prior to the experiment, participants were given detailed instructions outlining the experimental procedure as well as how to understand and interact with the user interface. Participants were asked to sign an integrity pledge after reading all of the instructions, stating that they would complete the experiment to the best of their abilities. After submitting their integrity pledge, participants were granted access to the experiment.

\subsubsection{Experimental Procedure}
Across all experiments, participants were randomly assigned 40 questions  (from the pool of 350 multiple-choice questions or the pool of 336 short-answer questions). The questions were sampled to balance across model confidence bins ensuring that all participants were exposed to questions at all levels of difficulty.

Furthermore, in Experiments 2a, 2b, and 2c, we balanced the types of explanation styles across questions so that each question was presented approximately the same number of times with each style. It should be noted that for each subject, each question was presented only once, and each question received only one explanation style. The counterbalancing, on the other hand, ensured that the same question had (roughly) an equal number of observations for each explanation style (across participants).

For the multiple choice questions, the participant's task was divided into two phases for each question (Extended Data Figure 2). In the first stage, participants had to provide a probability estimate that the LLM's answer was correct. In the second phase, participants had to answer the question with the assistance of the LLM. Participants were instructed to use their own knowledge as well as the LLM's response when making their own answer selection for this phase. For the short-answer questions, participants only had to provide a probability estimate that the LLM's answer was correct. They were instructed not to look up the answer. For the short-answer questions, the question answering phase was omitted to avoid participants from looking up the answers. 

At the end of the experiments, we administered a brief survey in which participants self-assessed their knowledge of topics associated with the multiple choice and short answer questions. Participants were asked to estimate the expected accuracy for each topic if they were presented with questions similar to those they encountered during the experiment.

\subsubsection{Creating explanation styles with varying degrees of uncertainty}
This section describes how we constructed prompts to elicit explanations with varying levels of uncertainty language in the experiments.  Table \ref{tab:explanationexamples} contains examples of explanations from the two experiments and styles. For additional details on the exact prompts used, see PT 3.

\begin{table}[h]
\tiny
  \caption{Example explanations for the answer by GPT-3.5 to a multiple-choice question used in Experiments 1a and 2a. The particular question is: \textit{A moving $1.5$ kg cart collides with and sticks to a $0.5$ kg cart which was initially at rest. Immediately after the collision, the carts each have the same \underline{\hspace{2cm}} as each other. Choose from the following options: A) Velocity B) Kinetic energy C) Mass D) Linear momentum}. The correct answer is A.}
  \label{tab:one_mcq_table}
 \begin{tabular}{p{0.28\linewidth}p{0.72\linewidth}}
 \toprule
    Prompt Style & Explanation\\
    \midrule
    \textbf{Experiment 1a} \\
    Baseline &
    The answer is [A] Velocity. This is because in an inelastic collision, the two objects involved will stick together and move with a common velocity after the collision. [$\ldots$] In this case, the initial momentum of the 1.5-kg cart is transferred to the combined mass of the two carts after the collision, resulting in the same velocity for both carts. Therefore, the correct answer is [A] Velocity. \\
    \midrule
    \textbf{Experiment 2a} \\
    Low Confidence \& Long &
    I am not completely sure that the answer is [A] because velocity is a vector quantity that includes both magnitude and direction. In this scenario, the carts may have the same magnitude of velocity, but they may not necessarily have the same direction. Therefore, I am not completely convinced that the answer is [A]. [$\ldots$] In this case, the carts would indeed have the same velocity as each other. \\
    \addlinespace
    Medium Confidence \& Long &
    I am somewhat sure that the answer is [A] because velocity is a vector quantity that takes into account both the speed and direction of an object's motion. In this scenario, the two carts are colliding and sticking together, [$\ldots$] Therefore, I am somewhat sure that the answer is [A] because the velocity of the two carts will be the same after the collision due to the conservation of momentum principle. \\
    \addlinespace
    High Confidence \& Long &
    I am sure that the answer is [A] because velocity is a vector quantity that describes the speed and direction of an object's motion. In this scenario, the two carts are colliding and sticking together, meaning they will have the same final velocity after the collision. [$\ldots$] This is also supported by the fact that the carts have different masses, so their kinetic energies and linear momenta will not be the same. Thus, the correct answer is [A] velocity. \\
    \addlinespace
    \addlinespace
    Low Confidence \& Short &
    I am not sure that the answer is [A] because the carts may have different velocities before and after the collision due to the conservation of momentum.\\
    \addlinespace
    Medium Confidence \& Short &
    I am somewhat sure that the answer is [A] because velocity is a vector quantity that takes into account both the speed and direction of an object's motion. In this scenario, the carts will have the same velocity after the collision because they are stuck together and moving in the same direction.\\
    \addlinespace
    High Confidence \& Short  &
    I am sure that the answer is [A] because velocity is a vector quantity that is conserved in collisions, meaning that the final velocity of the two carts will be the same. \\
    \addlinespace
    \addlinespace
    Low Confidence \& Uncertainty Only &
    I am not sure that the answer is [A].\\
    \addlinespace
    Medium Confidence \& Uncertainty Only &
    I am somewhat sure that the answer is [A].\\
    \addlinespace
    High Confidence \& Uncertainty Only  &
    I am sure that the answer is [A].\\
    \bottomrule
  \end{tabular}
  \label{tab:explanationexamples}
\end{table}

\paragraph{Experiment 1: baseline explanations.} To generate the baseline explanations for Experiment 1, we used a simple prompt that asked for an explanation, which would also repeat the answer.

\paragraph{Experiment 2: modified explanations.} In Experiment 2, explanations were manipulated in terms of the level of confidence expressed in the answer as well as the length of the answer. In total, the experiment included nine types of explanations (three levels of uncertainty x three levels of length). The three levels of confidence (low, medium, and high) were generated by prompts that instructed the LLM to ``mention you are not sure/somewhat sure/sure'' in the explanations respectively. The prompts elicited responses in which the beginning of each explanation indicated the level of uncertainty (e.g., ``I am not sure the answer is [B] because'' for the low confidence prompt). Note that expressions of uncertainty were not limited to the start of the explanation. Answers often contained additional explanations for why the LLM lacked confidence (e.g., ``further research may be required to confirm this,'' ``it is not possible to definitively state that...''). Experiment 2 also varied the length of the explanation across three levels: long, short, and uncertainty only. The long explanations were generated by not including any instruction regarding the length of the answer. The short explanations were generated by adding an instruction to use as few words as possible in the explanation. The uncertainty only explanation were generated by removing the rationale for the answer and included only the expression of uncertainty and the answer (e.g., ``I am not sure the answer is [B]''). 

For Experiment 2, the median lengths of the long and short explanations were: 115 and 34 words (GPT-3.5, Multiple Choice), 64 and 24 words (PaLM2, Multiple Choice) and 95 and 24 words (GPT-4o, Short-Answer). In comparison, the uncertainty only responses contained a median of 9 words across all variants of Experiment.

\subsection{Metrics}
\label{sec:metrics}
To investigate the relationship between the accuracy of answers to the multiple-choice and short-answer questions and the confidence (either human confidence or model confidence) associated with them, we utilize a range of metrics to evaluate this association. The primary focus is on understanding how well confidence levels correlate with the correctness of answers. To achieve this, we use both Expected Calibration Error (ECE) and the Area under the Curve (AUC) metric. These metrics assess the extent of overconfidence in predictions as well as the diagnostic effectiveness of confidence scores in distinguishing between correct and incorrect answers \cite{xiong2023can,tian2023just,jiang2021can, kadavath2022language, xiao2022uncertainty}. The use of AUC in this context parallels various metrics in psychology for metacognitive discrimination or sensitivity, which similarly aim to evaluate the effectiveness of confidence scores in distinguishing between correct and incorrect answers \cite{fleming2014measure}. In addition, in the Supplementary Information (``Overconfidence Error''), we also show results for the additional metric of Overconfidence Error (OE).

\subsubsection{Expected Calibration Error (ECE)}
We evaluate miscalibration using the Expected Calibration Error (ECE), as detailed in \cite{guo2017calibration,naeini2015obtaining}. ECE is calculated by averaging the absolute differences between accuracy and confidence across $M$ equal-width probability bins:

\begin{equation} 
\label{eqn:ece}
   \mathrm{ECE} = \sum_{m=1}^{M} \frac{|B_m|}{N} |conf(B_m) - acc(B_m)|
\end{equation}

where $N$ represents the total sample count, $B_m$ the $m$th confidence bin, and $acc(B_m)$ and $conf(B_m)$ denote the accuracy and average confidence for samples in the $m$th bin. ECE does not account for the direction of deviations between accuracy and confidence per bin respectively, so a nonzero ECE can indicate a mix of over- and underconfidence While recent work \cite{kumar2019verified,gruber2022better} has shown that ECE can under-estimate the true calibration error, the potential for under-estimation should not be a significant issue given that we are interested in analyzing differences in ECE  rather than  unbiased estimates of the error itself.

\subsubsection{Area under the Curve (AUC)} The AUC metric is employed to assess the diagnostic ability of confidence scores in distinguishing between correct and incorrect answers. Utilizing the Mann-Whitney U statistic approach, the AUC represents the probability that a randomly chosen correct answer has a higher confidence score compared to a randomly chosen incorrect answer:

\begin{equation} 
\label{eqn:auc}
   \mathrm{AUC} = \frac{1}{N_{pos} \times N_{neg}} \sum_{i=1}^{N_{pos}} \sum_{j=1}^{N_{neg}} I( C_i > C_j )  
\end{equation}

In this equation, $N_{pos}$ and $N_{neg}$ denote the counts of correct (positive) and incorrect (negative) answers, respectively. $C_i$ and $C_j$ represent the confidence scores of the $i$th and $j$th correct and incorrect answers, respectively. $I$ is an indicator function, which equals 1 if $C_i > C_j$ and 0 otherwise. This method evaluates each pair of correct and incorrect answers to determine if the confidence score for the correct answer surpasses that of the incorrect one. The AUC is then the fraction of these pairs satisfying this criterion, measuring the capability of confidence scores to differentiate between correct and incorrect responses, with AUC values ranging from 0.5 (indicating no better than chance discrimination) to 1 (signifying perfect discrimination). 

\subsection{Statistical Analysis}
To assess statistical significance, we utilize Bayes factors (\emph{BF}s) to determine the extent to which the observed data adjust our belief in the alternative and null hypotheses. Values of 3 $<$ \emph{BF} $<$ 10 and \emph{BF} $>$ 10 indicate moderate and strong evidence against the null hypothesis, respectively. Similarly, values of 1/10 $<$ \emph{BF} $<$ 1/3 and \emph{BF} $<$ 1/10 indicate moderate and strong evidence in favor of the null hypothesis, respectively. We report Bayes factors for Bayesian t-tests using the default priors as recommended by \cite{rouder2012default}.

\section{Results}
%\subsection*{Default explanations do not enable humans to accurately assess what LLMs know}
We start by examining the results from Experiment 1 and compare human and model confidence in the case where LLMs generate default explanations for participants. We present the results for two different metrics: 1) Expected Calibration Error (ECE), which assesses the degree to which confidence scores from the model or the human reflect the true accuracy of the LLM, and 2) Area Under the Curve (AUC), which assesses the degree to which confidence scores discriminate between correct and incorrect responses (see Methods for details). The findings indicate that there is a significant gap, as measured by calibration and discrimination, between what LLMs know and what humans believe they know based on default explanations.

\begin{figure}[h!]
\centering
% left, bottom, right, top
\includegraphics[width=1.0\linewidth,trim=0cm 0cm 0cm 0cm, clip]{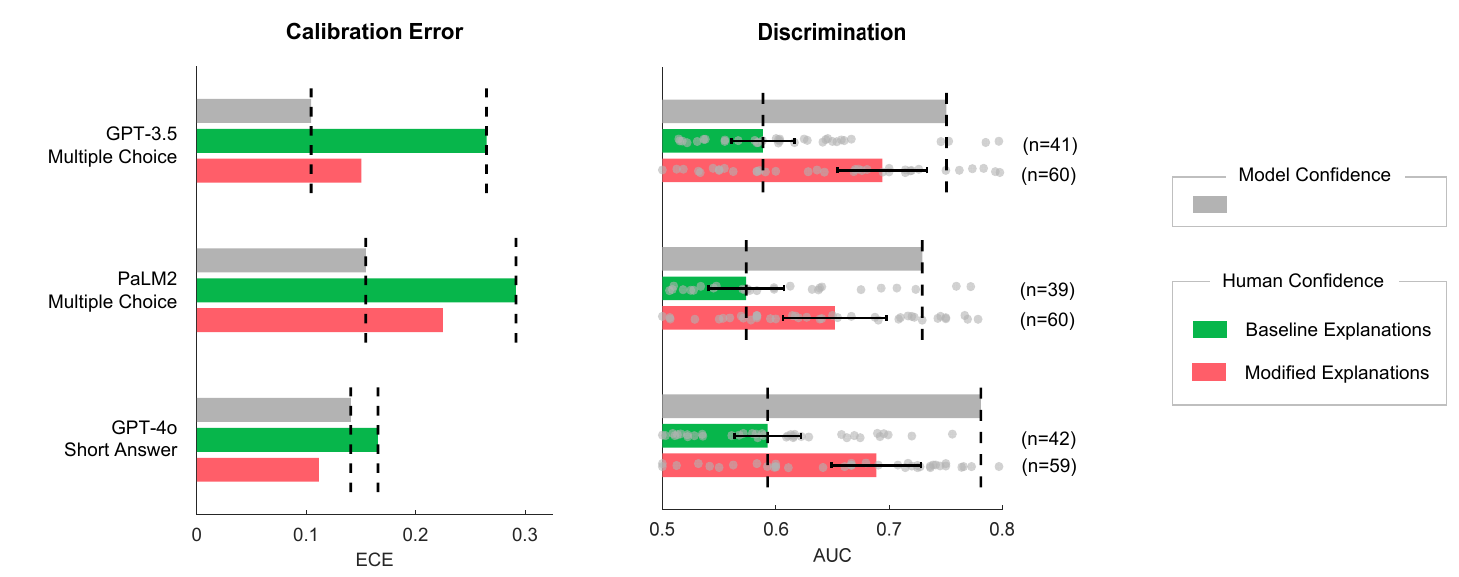}
\caption{Calibration error and discrimination for model confidence and human confidence across the behavioral experiments and LLMs. Calibration error is assessed by ECE (lower is better) while discrimination is assessed by AUC (higher is better). Vertical dashed lines represent the calibration and discrimination gap between model confidence and human confidence for unmodified explanations (Experiments 1a, 1b, and 1c). For human confidence, data points represent the AUC values computed separately for each participant ($n$ shown in figure), and error bars represent the 95\% confidence interval of the mean across participants. Because of data sparsity, the ECE values were computed at the group level.}
\label{fig:allresults}
\end{figure}

\subsubsection{Calibration gap}
Figure \ref{fig:allresults}, left panel, shows the ECE for both model and human confidence. The results show a calibration gap; across the different types of LLMs and types of questions (multiple choice and short answer), the ECE metric is much lower for model confidence (in gray) than for human confidence with baseline explanations (in green).  This gap demonstrates that standard explanations provided by the LLM do not enable participants to judge the likelihood of correctness of the LLM's answers, leading to a misalignment between perceived accuracy and actual LLM accuracy.

\begin{figure}[h!]
% left, bottom, right, top
\includegraphics[width=1.0\linewidth,trim=2cm 3cm 7cm 0cm, clip]{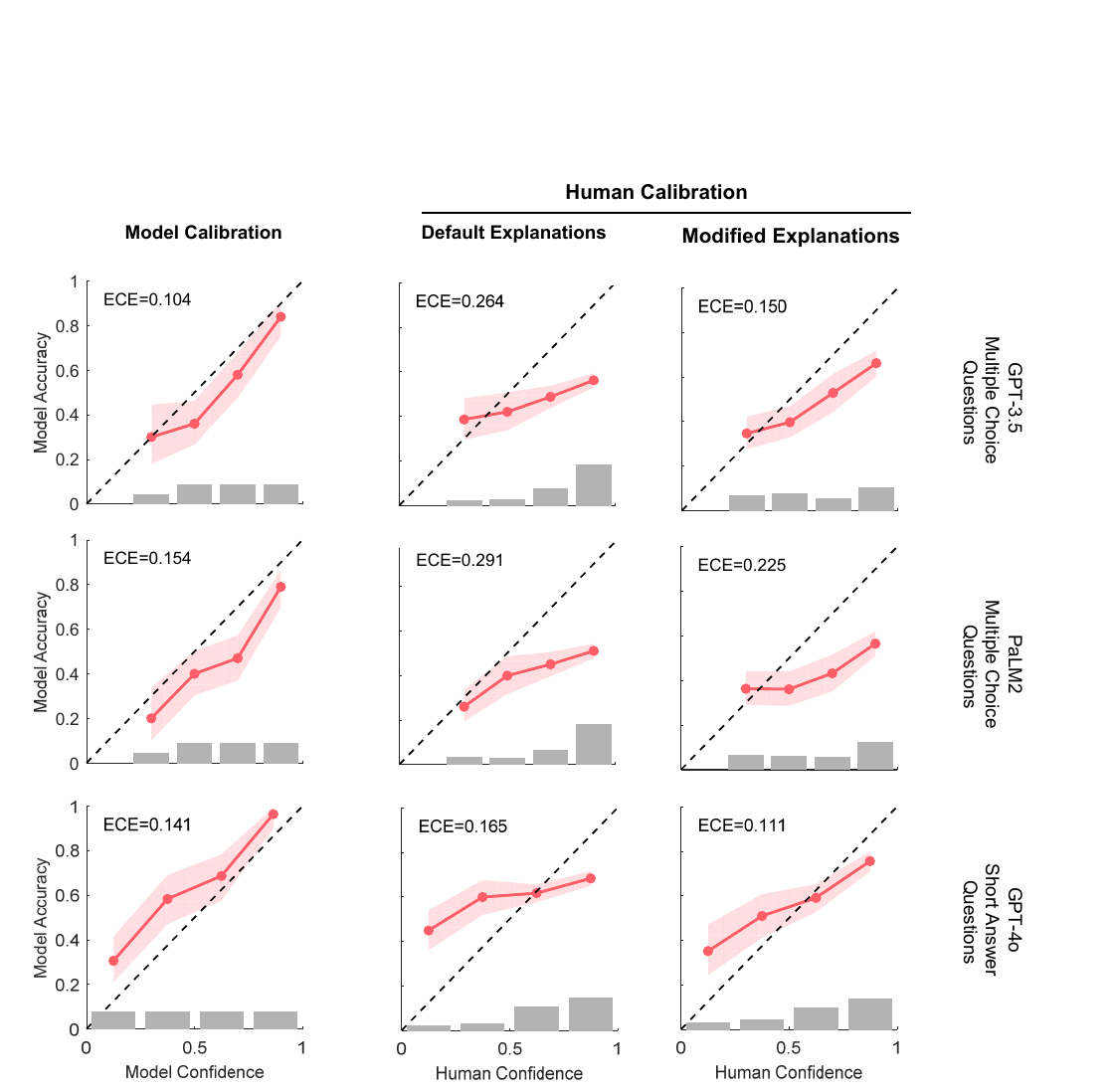}
\caption{Calibration diagrams for model confidence and human confidence across Experiments 1 and 2. The top and middle rows show results for multiple-choice questions with the GPT-3.5 and PaLM2 models, respectively. The bottom row shows results for short-answer questions with the GPT-4o model. The histograms at the bottom of each plot show the proportion of observations in each confidence bin (values are scaled by 30\% for visual clarity). Shaded regions represent the 95\% confidence interval of the mean computed across participants and questions.}
\label{fig:calibrationresults}
\end{figure}

Figure \ref{fig:calibrationresults} expands on the calibration results in Figure \ref{fig:allresults} to show detailed calibration results for each LLM and each experimental condition. The diagrams show how well model confidence (left column) and human confidence (right two columns) are calibrated. The ideal calibration (i.e., ECE=0) would yield results along the diagonal. For multiple-choice questions, both LLMs have a tendency to be overconfident, resulting in calibration lines below the diagonal. For the short-answer questions, the LLM is somewhat underconfident. Comparing the LLM to the human calibration in Experiment 1 (middle column), the results show that for the multiple choice questions, human miscalibration is primarily due to overconfidence, indicating that people generally believe that LLMs are more accurate than they actually are. The histograms (inset panels) demonstrate that a significant portion of the calibration error is due to participants' propensity to produce high confidence scores, even though the model accuracy for the associated questions is much lower than expected based on confidence.

\subsubsection{Discrimination gap}
Participants are not very good, relative to the LLM, at discriminating between which answers are likely correct or incorrect based on the default explanation. We assess discrimination using the AUC metric applied to the human confidence ratings. Figure \ref{fig:allresults}, right panel, shows the AUC for both model and human confidence. The results show a gap between how well model and human confidence discriminate between correct and incorrect answers. The LLM model confidence discriminates between correct and incorrect answers well above chance (GPT-3.5 AUC=.751, PaLM2 AUC=0.746 for the multiple choice questions and GPT-4o AUC=0.781 for the short-answer questions). In contrast, participants who view the default explanations in Experiment 1 were only slightly better than random guessing (AUC=0.589 and AUC=0.602 for the multiple choice explanations by GPT-3.5 and PaLM2 respectively and AUC=0.592 for the short-answer explanations by GPT-4o). Therefore, default explanations lead to a discrimination gap as well. 

\subsection{Explanation style and length affect human confidence}
Experiment 2 evaluates how human confidence is affected by the degree of uncertainty expressed in LLM explanations (across three levels of confidence) as well as the overall length of the LLM explanation (across three levels of length). 

\begin{figure}[h!]
% left, bottom, right, top
\centering
\includegraphics[width=1.0\linewidth,trim=0cm 7cm 0cm 4cm, clip]{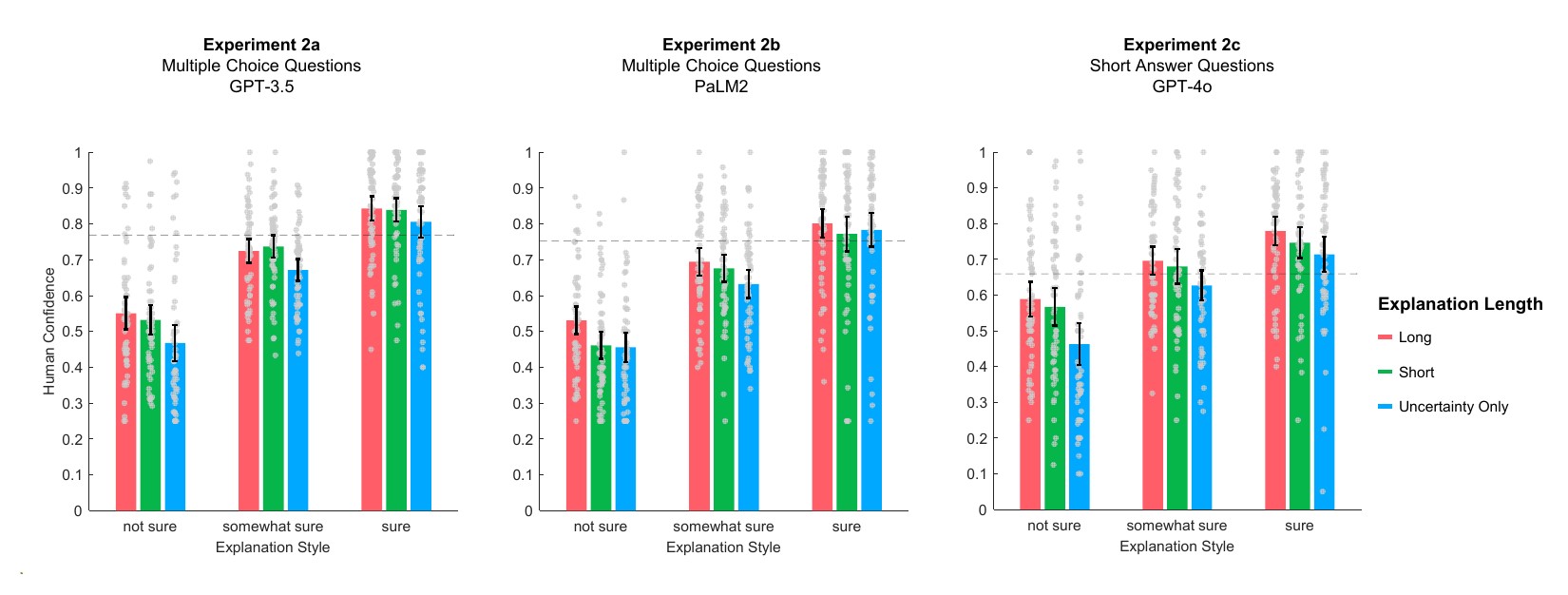}
\caption{Mean human confidence across LLM explanation styles varying in uncertainty language and length. Data are presented as mean values of participant confidence in Experiments 2a ($n$=60), 2b ($n$=60), and 2c ($n$=59). For reference, dashed lines show the average human confidence for the baseline explanations in Experiment 1a, 1b, and 1c. Error bars represent the 95\% confidence interval of the mean across participants.}
\label{fig:styles}
\end{figure}

Figure \ref{fig:styles} shows that the type of uncertainty language used in the explanations has a strong influence on human confidence regardless of the type of LLM that produced the explanation or the type of question. Low confidence explanations (``I am not sure'') produced significantly lower human confidence than medium confidence explanations (``I am somewhat sure''); $BF>100$ across Experiments 2a, 2b and 2c. Similarly, medium confidence explanations produced lower human confidence than high confidence explanations; $BF>100$ across Experiments 2a, 2b, and 2c. The Supplementary Information (``Human Confidence Agreement'') shows an analysis of the reliability of the confidence ratings across participants.

In addition, the length of the explanations also affected the human confidence in the LLM answers. Long explanations led to significantly higher confidence than the short explanations ($BF=25$ with data combined across Experiments 2a, 2b and 2c) and short explanations led to significantly higher confidence than the responses that only contained the uncertainty expression ($BF>100$ with data combined across Experiments 2a, 2b, and 2c). The additional information presented in longer explanations did not enable participants to better discriminate between likely correct and incorrect answers for longer explanations. Across Experiments 2a, 2b, and 2c, the mean participant AUC is 0.54 and 0.57 for long and uncertainty-only explanations, respectively ($BF=.23$). Therefore, the length of the answer led to an increase in human confidence without any corresponding change in sensitivity to discriminating between correct and incorrect answers. 

The results confirm that people can appropriately interpret verbal cues about uncertainty and that manipulating the length of the explanation can directly impact human confidence.

%\subsection*{Model-confidence modified explanations reduce the calibration and discrimination gaps}
\subsection{Reducing the calibration and discrimination gap}
Having established in Experiment 2 that the uncertainty language in the LLM explanation can modify human confidence, we now evaluate whether linking the type of uncertainty language to the LLM model confidence (i.e., showing a low/medium/high confidence explanation when model confidence is low/medium/high) can reduce the calibration and discrimination gap. 

\subsubsection{Selecting explanations based on model confidence}
\label{sec:selectionmethodsection}
We simulated the effect of aligning the explanation style to model confidence by a simple decision rule. With this rule, we select the type of explanation $s \in $ \{low confidence, medium confidence, high confidence\} based on the LLM model confidence score $p$: 
\begin{equation}
\begin{aligned}
  s &=
  \begin{cases}
   \mathrm{low \; confidence} & \text{if } p \leq \theta_1 \\
   \mathrm{medium \; confidence} & \text{if } \theta_1 < p \leq \theta_2\\ 
   \mathrm{high \; confidence} & \text{if } \theta_2 < p\\ 
  \end{cases}
\end{aligned}
\label{eqn:selectionrule}
\end{equation}
The parameters $\theta_1$ and $\theta_2$ determine the ranges where low, medium, and high confidence explanations are chosen. The application of this rule to a given parameter setting leads to a  participant's estimates being filtered out if the explanation style used for a specific question does not match the selected style. This allowed us to simulate the effect of participants receiving different types of explanations based on model confidence (i.e., lower confidence explanations for low model confidence and high confidence explanations for high model confidence). The Supplementary Information (``Optimization Procedure'') provides details on the optimization procedure and also a demonstration that the results are not particularly sensitive to the parameter settings.

\subsubsection{Calibration and Discrimination Results}
Figure \ref{fig:allresults} shows the calibration and discrimination results when the selection rule is applied to the results from Experiment 2. The results in Figure \ref{fig:allresults} (left panel, red bars) show that the calibration gap has narrowed substantially. While there is still generally a higher calibration error for human confidence relative to model confidence, the calibration gap has decreased for all three LLMs relative to the baseline explanations in Experiment 1. Furthermore, Figure \ref{fig:allresults} (right panel) shows that the discrimination gap (as measured by AUC) has also been narrowed relative to the baseline explanations across LLMs and question types ($BF>100$, $BF=6.48$, and $BF>100$ for Experiments 2a, 2b, and 2c, respectively). Therefore, the results show that selecting the type of explanation based on LLM model confidence improves both calibration and discrimination performance, as human confidence in the LLM becomes more closely aligned with the LLM's actual accuracy.

\subsection{Accuracy}
\subsubsection{Participants lack specialized knowledge}
For the experiments with multiple choice questions (1a, 1b, 2a, and 2b), participants provided their own answer after seeing the answer from the LLM. This allowed us to analyze whether participants have any independent knowledge from the LLM that allowed them to improve on LLM accuracy. In experiments 1a and 2a with GPT-3.5, participants' average answer accuracy was 51\%, closely aligning with LLM's 52\% accuracy rate. Similarly, for the multiple choice experiments 1b and 2b with PaLM2, average participant accuracy was 45\%, similar to the 47\% accuracy rate for the LLM. In the majority (82\%) of responses across all multiple choice experiments, participants selected the response that agreed with the LLM's explanation.

When participants chose to alter the answer, the average accuracy was 33\% which is lower than the LLM's accuracy of 39\% for these particular questions. These findings suggest limited success in participants' ability to accurately answer the questions independent of the LLM's explanation. This is consistent with findings from \cite{hendryckstest2021}, showing that Mechanical Turk workers without specialized knowledge (akin to our participant pool) scored 35\% accuracy on similar questions.

When we applied the selection rule and the explanation type was aligned with model confidence, human decision accuracy in Experiments 2a and 2b did not improve for the selected questions (even though discrimination and calibration improved). This shows that accurate uncertainty communication by the LLM allowed participants to recognize when the LLM was providing a likely correct or incorrect answer, but the lack of accurate human knowledge independent from the LLM prevented participants from improving on the LLM answer.  

\subsubsection{Self-assessed expertise does not affect performance}
At the end of the experiment, participants estimated the performance they would achieve on similar questions for each of the 10 topics in the sample of MMLU questions. The median of these self-assessed expertise estimates did not substantially vary between topics: from 30\% (e.g., high school physics) to 45\% (e.g., high school world history). Examining the impact of perceived expertise on accuracy estimation, we divided participants into two groups based on whether their self-rated expertise was above or below 50\% separately for each of the 10 topics. For the experiments with GPT-3.5, the higher expertise groups generally had better discrimination (AUC 0.600 vs. AUC 0.579), but there was no evidence that this difference was significant (BF $<$ 1). In addition, the calibration error was comparable between the two groups (ECE =.289 vs. .292). Similarly, no effect of expertise was found for the experiments with PaLM2. Therefore, participants who considered themselves more knowledgeable about a topic were not more adept at estimating the LLM's performance in that area.

\section{Discussion}
Our research focused on bridging the gap between what an LLM knows and what users perceive it knows. This gap is critical, especially as reliance on LLMs for decision-making across various domains is rapidly increasing. 

Research on LLMs has begun to address these challenges, with a focus on improving uncertainty communication and the quality of explanations. Several studies have explored LLM confidence in answering multiple-choice questions, focusing on how well the models' self-reported confidence aligns with their actual accuracy \cite{kadavath2022language, openai2023gpt4, hendryckstest2021, xiong2023can} and whether users can accurately assess the reliability of the explanations provided \cite{tanneru2023quantifying}. The work by \cite{zhou2024relying} investigates how users respond to verbal phrases of uncertainty in a simulated trivia task but does not employ actual LLM outputs. Overall, there has been little research examining user confidence in LLM output. Our work uses actual LLM outputs and its confidence in an attempt to quantify the calibration and discrimination gap. As a result, we directly address the issue of miscommunication of uncertainty from LLMs to humans.

Our results showed that users consistently overestimated how accurate LLM outputs were, especially when they relied on the models' default explanations. This was true for three different LLMs and two different types of questions (multiple choice and short answer). This tendency toward overconfidence in LLM capabilities is a significant concern, particularly in scenarios where critical decisions rely on LLM-generated information. The inability of users to discern the reliability of LLM responses not only undermines the utility of these models but also poses risks in situations where user understanding of model accuracy is critical. 

In addition, the results also showed a length bias where longer explanations led to higher human confidence levels even though they did not contain any additional information to help users to better discriminate between likely correct and incorrect answers. This suggests that users were processing the explanations at a shallow level, relying on simple textual cues such as overall length to predict LLM accuracy. This result is consistent with studies in social psychology and communication research which suggest that longer answers or explanations may be perceived as more persuasive or credible, even when they do not contain more meaningful information \cite{petty1984effects, oppenheimer2006consequences}. This length bias has also been found in domains such as peer reviews, where longer reviews are perceived as more persuasive and informative even if the information content remains the same \cite{goldberg2023peer}.

Although default LLM explanations do not enable users to perceive what the models truly know, this research shows that a simple approach based on tailored explanations can bridge this perception gap. This was achieved by altering the prompts used to generate explanations based on model confidence, allowing for better control over how uncertainty was expressed in the responses. Specifically, we designed these prompts to induce varying degrees of certainty in the explanations, ranging from expressions of low confidence (e.g., "I am not sure the answer is [B] because") to high confidence (e.g., "I am confident the answer is [B] because"). By modifying the language of the LLM’s responses to better reflect model confidence, users showed improved calibration in their assessment of the LLM’s reliability and were better able to discriminate between correct and incorrect answers. This improvement underscores the importance of transparent communication from LLMs, suggesting a need for researchers to investigate how model explanations affect user perception.

One limitation of the current study is the focus on a specific type of question involving a small number of response alternatives (multiple choice) and short-answers to open-ended questions. The extent to which these results apply to longer open-ended questions remains an open question. Further research could investigate the applicability of our findings across a broader range of scenarios. Another limitation of this study is that our approach to modifying the prompt based on internal uncertainty required the LLM to be prompted twice: once to read out the answer and model confidence, and again to produce an explanation modified by the model confidence. Future research could investigate how to produce confidence-modified explanations in a single step.

Another important area for future research is to understand the fundamental causes of the miscommunication of uncertainty. %How is it possible that the
Why do LLMs generate calibrated model confidences while also producing explanations that are not consistent with those confidences? One hypothesis is that current LLMs are aligned to human preferences using reinforcement learning from human feedback (RLHF, \cite{ouyang2022training}, which produces some built-in biases. In these RLHF procedures, various types of explanations are presented to human participants, who can then choose their preferred explanations. LLMs are then fine-tuned based on human preference data, making them more likely to produce explanations that people prefer. While RLHF encourages human-aligned output, it inevitably reproduces any human preference biases. For example, people prefer detailed and generally longer explanations \cite{bower2024experts, saito2023verbosity}. As a result, LLMs trained on these human preferences may produce explanations that are overly convincing, potentially misleading users about the reliability of the information. 

An alternative hypothesis to the production of overconfident explanations lies in the autoregressive nature of well-established LLMs. In particular, we conjecture that after committing to an answer (encoded as a sequence of tokens), the model will generate a sequence of tokens (explanation) that maximizes the likelihood of the previous answer, effectively resulting in an assertive answer. A similar hypothesis was also presented in \cite{azaria2023internal}. Interestingly, the possibility that the LLM's choice of a particular answer inflates the rationale for that answer resembles the phenomenon of choice-supportive biases in psychology \cite {mather2000misremembrance}. After making a decision, people tend to overestimate the desirability of the chosen option while underestimating the desirability of rejected alternatives. This can make them feel more confident in their decision than they were when they first made it.

Our work shares some parallels with prior studies on the human perception and evaluation of AI-generated explanations in the domain of machine learning classifiers (see \cite{Rongetal2023} for an overview). These studies frequently employ feature highlighting to explain what areas of the image \cite{Smith-Renner2020} or what fragments of documents \cite{feng2019can} can support the suggested classification. Studies have found mixed evidence for the effectiveness of these types of AI explanations in human decision-making  \cite{steyverskumar2023, bansal2021does,buccinca2021trust,wangetal2022}. These results highlight the challenge of ensuring that AI-generated explanations align with human expectations and allow humans to distinguish between correct and incorrect answers. 
 
In conclusion, our research highlights the critical role of clear and accurate communication in the interaction between users and LLMs. Enhancing the alignment between model confidence and the user's perception of model confidence can lead to more responsible and trustworthy use of LLMs, particularly in areas where the accuracy of AI-generated information is critical.

\section*{Data Availability}
All behavioral data as well as data produced by the Large Language Models used in this study are publicly available from the following OSF repository: \url{https://osf.io/y7pr6/} %(\cite{steyverscode})

\section*{Code Availability}
The code used for data analysis and extracting LLM model confidence is available from the following OSF repository: \url{https://osf.io/y7pr6/} %(\cite{steyverscode})

\acks{This research was supported by NSF under award 1900644 (P.S., and M.S.)}

% Manual newpage inserted to improve layout of sample file - not
% needed in general before appendices/bibliography.
\newpage
\bibliography{paper}
\newpage

\appendix
\section{Additional Experiment Results}

\subsection{Demographic Information by Experiment}
%TC:ignore
\begin{table}[h!]
  \caption{Breakdown of demographic information by experiment. Note that Sex was not reported by all participants.}
  \label{tab:freq}
  \centering
  \begin{tabular}{lllllll}
    \toprule
   Experiment & \#Participants & \#Male & \#Female & Median Age & Min Age & Max Age \\
    \midrule
    1a & 41 & 18 & 20 & 35 & 21 & 71 \\
    1b & 39 & 21 & 18 & 36 & 21 & 69 \\
    1c & 42 & 18 & 19 & 32 & 18 & 69 \\
    2a & 60 & 31 & 28 & 35 & 18 & 79 \\
    2b & 60 & 28 & 32 & 36 & 19 & 62 \\
    2c & 59 & 22 & 29 & 33 & 18 & 74 \\
  \bottomrule
\end{tabular}
\label{tab:demogr}
\end{table}

%            whexp             GroupCount    median_Age    min_Age    max_Age
%    ______________________    __________    __________    _______    _______
%    {'exp1a_GPT_alldata' }        38            35          21         71   
%    {'exp1b_PALM_alldata'}        39            36          21         69   
%    {'exp1c_GPT_alldata' }        37            32          18         69   
%    {'exp2a_GPT_alldata' }        60            35          18         79   
%    {'exp2b_PALM_alldata'}        60            36          19         62   
%    {'exp2c_GPT_alldata' }        53            33          18         74   
    
%            whexp                   Sex           GroupCount
%    ______________________    ________________    __________
%    {'exp1a_GPT_alldata' }    {'Female'      }        20    
%    {'exp1a_GPT_alldata' }    {'Male'        }        18    
%    {'exp1b_PALM_alldata'}    {'Female'      }        18    
%    {'exp1b_PALM_alldata'}    {'Male'        }        21    
%    {'exp1c_GPT_alldata' }    {'Female'      }        19    
%    {'exp1c_GPT_alldata' }    {'Male'        }        18    
%    {'exp2a_GPT_alldata' }    {'DATA_EXPIRED'}         1    
%    {'exp2a_GPT_alldata' }    {'Female'      }        28    
%    {'exp2a_GPT_alldata' }    {'Male'        }        31    
%    {'exp2b_PALM_alldata'}    {'Female'      }        32    
%    {'exp2b_PALM_alldata'}    {'Male'        }        28    
%    {'exp2c_GPT_alldata' }    {'DATA_EXPIRED'}         2    
%    {'exp2c_GPT_alldata' }    {'Female'      }        29    
%    {'exp2c_GPT_alldata' }    {'Male'        }        22 

\subsection{Experiment 3} 
We conducted an additional experiment that used a different prompting method to modify the uncertainty language expressed in explanations for multiple choice questions. In this approach, we gave the LLM the baseline explanation from Experiments 1a and 1b in the prompt and instructed the LLM to ``rewrite the explanation as if you are {\it not sure/somewhat sure/sure}'' for each of the three confidence levels. In contrast to the prompts for Experiments 2, this prompt leads to less stereotyped expressions of uncertainty within each response. This experiment allowed us to investigate whether the results generalize across different approaches to express uncertainty in the explanations.

In Experiment 3a and 3b, we used GPT-3.5 and PaLM2 respectively to generate the explanations mirroring the experiments 2a and 2b. The methodology was the same as in Experiment 2a and 2b except that we did not include a length manipulation. The experiment was conducted with 81 participants (40 in Experiment 3a, 41 in Experiment 3b). 

The results of Experiment 3 are shown in Supplementary Table \ref{tab:addresults} along with the results from Experiment 1 and 2. The pattern of results is the same as in Experiments 2a and 2b. Human confidence exhibited smaller calibration error (ECE) and larger discrimination (AUC) relative to the baseline results in Experiments 1a and 1b. These results show that the experimental results from 2a and 2b generalize to different prompts to elicit uncertainty. 

\begin{table}[h!]
  \caption{Expected Calibration error (ECE), Overconfidence Error (OE), and Area under the Curve (AUC) of model and human confidence across experiments.}
  \label{tab:freq}
  \begin{tabular}{lllll}
    \toprule
    LLM & Confidence Score&ECE&OE&AUC\\
    \midrule
    GPT-3.5\\ 
    & model confidence                                     & .104 & .064 & .751 \\
    & human confidence: experiment 1a, default explanations & .264 & .220 & .589 \\
    & human confidence: experiment 2a, modified explanations & .150 & .107 & .694 \\ % & .183 & .145 & .692 (old experiment)
    & human confidence: experiment 3a, modified explanations & .158 & .127 & .678 \\ % .158 & .121 & .678
    PaLM2\\
    & model confidence                                     & .154 & .098 & .746 \\
    & human confidence: experiment 1b, default explanations & .291 & .229 & .602 \\
    & human confidence: experiment 2b, modified explanations & .225 & .168 & .652 \\ % & .134 & .078 & .655 (old experiment)
    & human confidence: experiment 3b, modified explanations & .195 & .155 & .689 \\ %  .195 & .155 & .689
    GPT-4o\\
    & model confidence                                     & .141 & .008 & .781 \\
    & human confidence: experiment 1c, default explanations & .165 & .084 & .593 \\
    & human confidence: experiment 2c, modified explanations & .111 & .008 & .689 \\ % & .134 & .078 & .655 (old experiment)
  \bottomrule
\end{tabular}
\label{tab:addresults}
\end{table}

\subsection{Overconfidence Error}
Supplementary Table \ref{tab:addresults} also includes results for an additional performance metric, the Overconfidence Error (OE). The metric is an adaptation of the ECE formula, specifically focusing on cases of overconfidence \cite{thulasidasan2019mixup}:

\begin{equation} 
\label{eqn:oe}
   \mathrm{OE} = \sum_{m=1}^{M} \frac{|B_m|}{N} [ conf(B_m) \times \mathrm{max}( 0, conf(B_m) - acc(B_m)) ]
\end{equation}

This penalizes predictions by the weight of the confidence but only when confidence exceeds accuracy. The results show that human confidence for default explanations leads to the largest overconfidence error and that the model confidence modified explanations lowers the overconfidence error.

\begin{figure}[h!]
  \centering
  \includegraphics[width=0.32\linewidth]{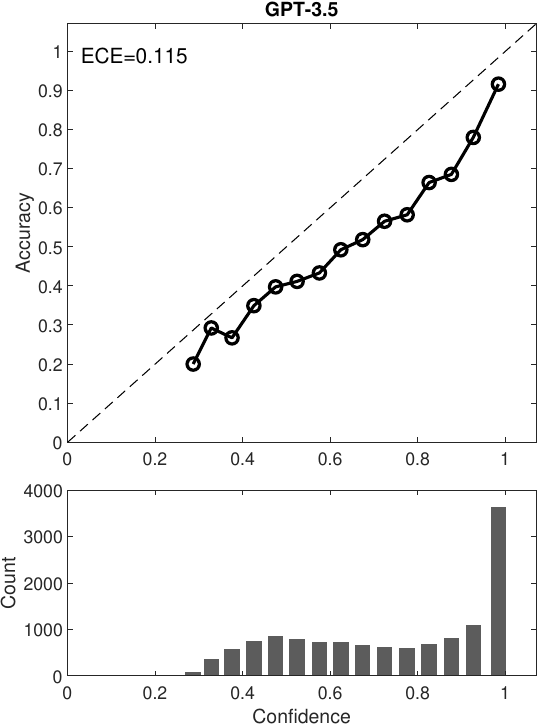}
  \includegraphics[width=0.32\linewidth]{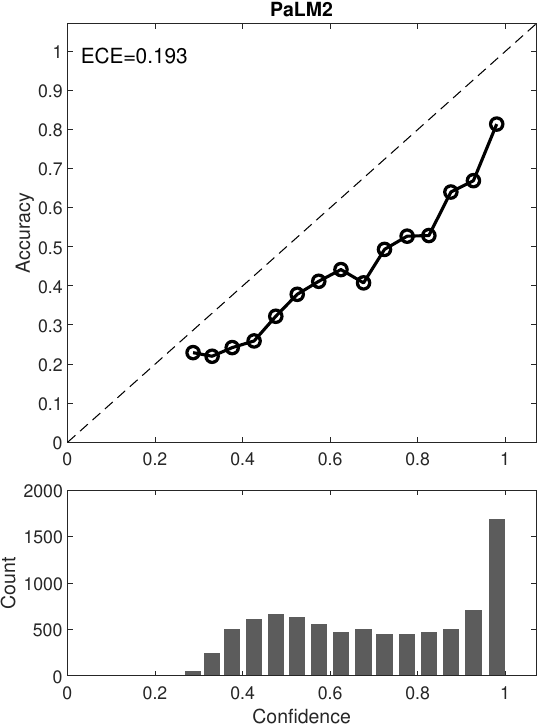}
  \includegraphics[width=0.32\linewidth]{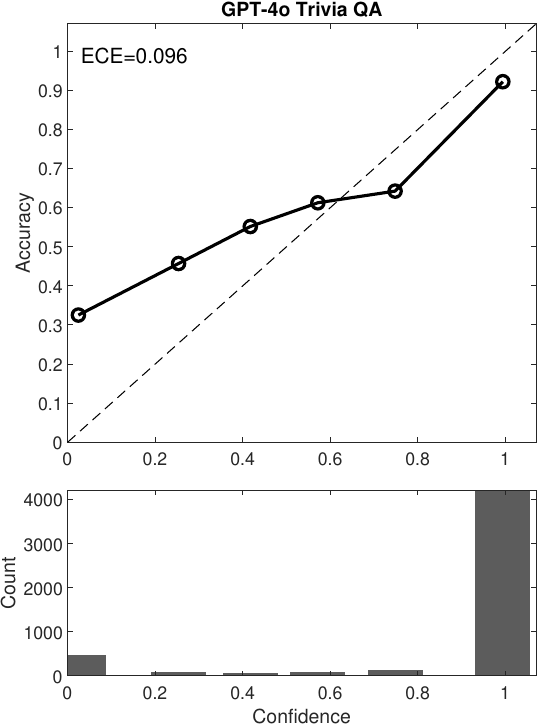}
  \caption{Calibration diagrams for the full set of MMLU questions for GPT-3.5 and PaLM2 (left and middle panel) and the 5000 question sample of the Trivia QA data set using the GPT-4o model (right panel).}
  %\caption{Calibration diagram for GPT-3.5 model for full set of MMLU questions.}
  \label{fig:calibrationfull}
\end{figure}

\subsection{Human Confidence Agreement}
For Experiment 2, we analyzed the degree to which there is agreement in the confidence ratings across participants. We assessed agreement with the mean correlation between any pair of participants. Because there were few cases where any pair of participants were rating the LLM response for the same question at the same level of uncertainty language at the same level of length, we conducted the analysis at the level of the experimental manipulations. With this approach, each participant was characterized with 9 confidence ratings corresponding to the mean rating for each of 3 uncertainty levels x 3 length levels. At this aggregate level, mean participant-to-participant correlation was 0.550, 0.547, and 0.336 for Experiments 2a, 2b, and 2c respectively. This shows that there was moderate agreement in human confidence about the differences in levels of uncertainty and length.

\begin{figure}[h!]
  \centering
  \includegraphics[width=0.32\linewidth]{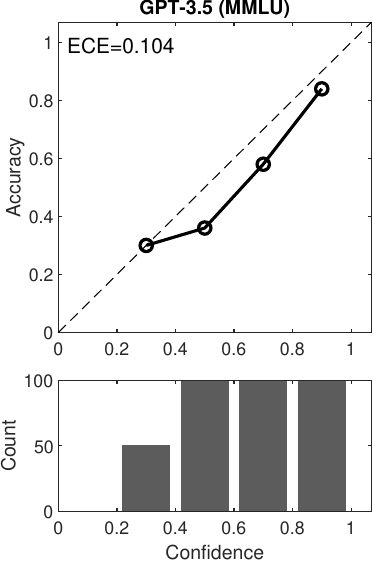}
  \includegraphics[width=0.32\linewidth]{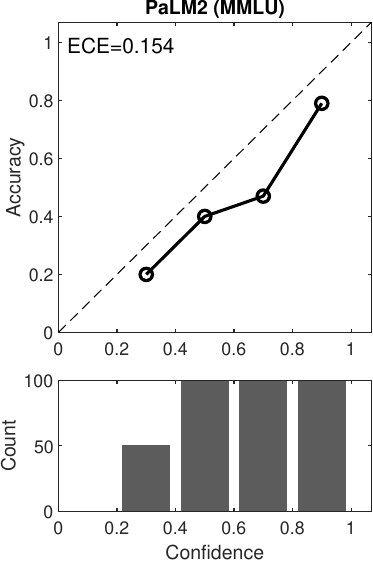}
  \includegraphics[width=0.32\linewidth]{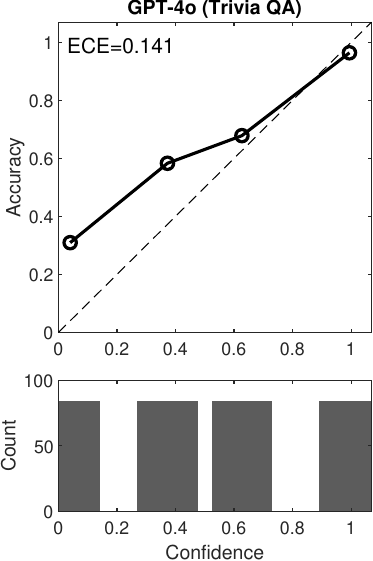}
  \caption{Calibration diagrams for the subsets of questions used for the behavioral experiments. Left and middle panels show the calibration for the GPT-3.5 and PaLM2 models for the MMLU multiple choice questions. The right panel shows the calibration for the GPT-4o model for the Trivia QA questions. Note that the lower count of multiple choice questions in the lowest confidence bin is due to the sparsity of questions in that confidence bin.}
  \label{fig:calibrationsubset}
\end{figure}

\newpage
 
\section{Additional Model Confidence Results}
Supplementary Figure \ref{fig:calibrationfull} shows the calibration diagrams for the full set of 14,042 test questions from the MMLU dataset and the 5000 questions from the Trivia QA dataset. For comparison, Supplementary Figure \ref{fig:calibrationsubset} shows the calibration diagrams for the subset of 350 multiple choice and 336 short-answer questions used for the behavioral experiments.

\paragraph{MMLU multiple choice questions.} For GPT-3.5, the accuracy across all 14,042 questions is 63\% with an AUC of 0.78. When computing model confidence, 8.7\% of the answers were incomplete and were removed from consideration. For PaLM2, the accuracy is 51\% with an AUC of 0.73. Furthermore, we confirmed that we could replicate the GPT-4 Technical Report's \cite{openai2023gpt4} five-shot results. Five-shot prompting with GPT-3.5 (detailed in Appendix 8 of the report) resulted in 71\% accuracy (compared to 70\% reported in Supplementary Table 2 of the report). The zero-shot approach is the focus of this paper. The zero-shot approach simplifies the construction of explanations, and our goal is not to maximize language model accuracy.
% five-shot: ECE of 0.078 and 0.007\% incomplete answers

\paragraph{Trivia QA short answer questions.} For GPT-4o, the accuracy across the 5000 question sample is 85\% with an AUC of 0.85. For the 336 question subset used for the behavioral experiments, accuracy is 63\% with an AUC of 0.78. The lower accuracy is caused by the uniform sampling across confidence bins that results in an over-representation of questions with lower confidence for which the model is also less accurate.

\newpage

\section{Optimization Procedure}
As discussed in the main text, we applied a selection rule to link the level of confidence, $s \in $ \{low confidence, medium confidence, high confidence\}, in the explanation to model confidence $p$:  
\begin{equation}
\begin{aligned}
  s &=
  \begin{cases}
   \mathrm{low \; confidence} & \text{if } p \leq \theta_1 \\
   \mathrm{medium \; confidence} & \text{if } \theta_1 < p \leq \theta_2\\ 
   \mathrm{high \; confidence} & \text{if } \theta_2 < p\\ 
  \end{cases}
\end{aligned}
\label{eqn:selectionrule}
\end{equation}
The parameters $\theta_1$ and $\theta_2$ determine the ranges where low, medium, and high confidence explanations are chosen. The application of this rule to a given parameter setting leads to any participant estimates being filtered out if the explanation style used for a specific question does not match the selected style. To apply the selection rule, we ignored the variations in length in Experiment 2. Therefore, for both Experiment 2 and 3, the selection rule considered a choice of one of three explanation styles for each question (i.e., low, medium, and high confidence explanations).

\begin{figure}[h!]
  \centering
  \includegraphics[width=0.32\linewidth]{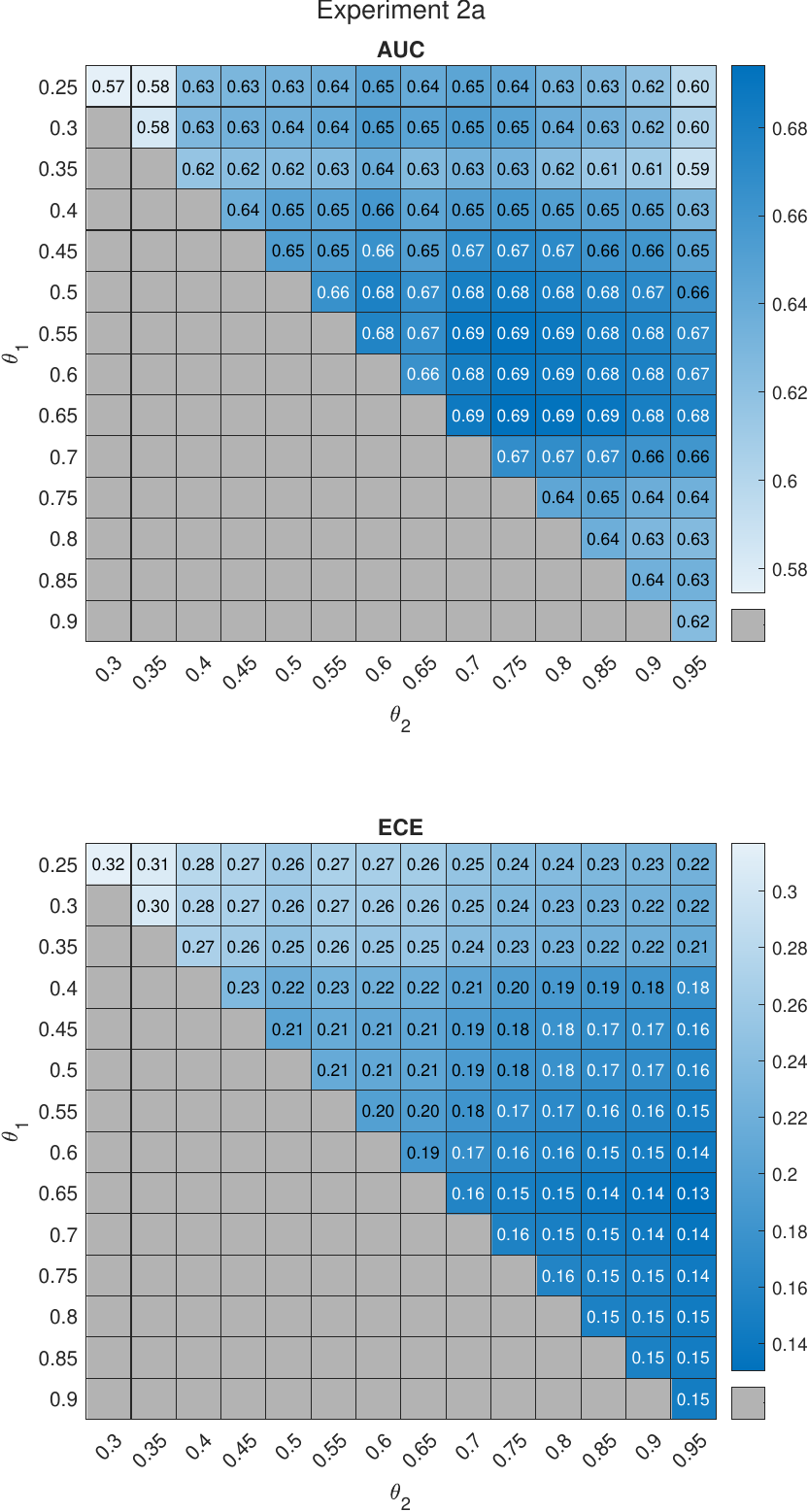}
  \includegraphics[width=0.32\linewidth]{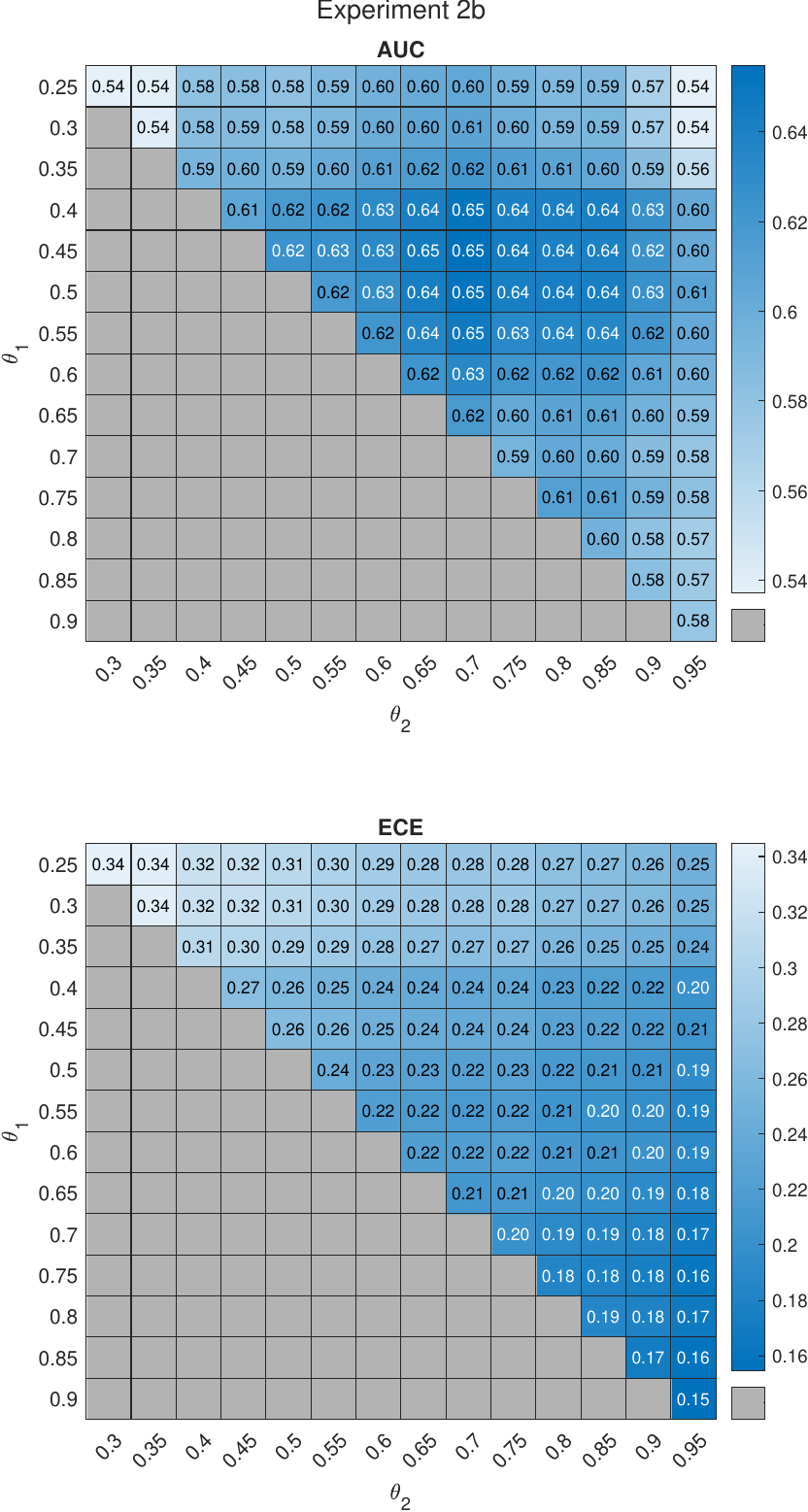}
  \includegraphics[width=0.345\linewidth]{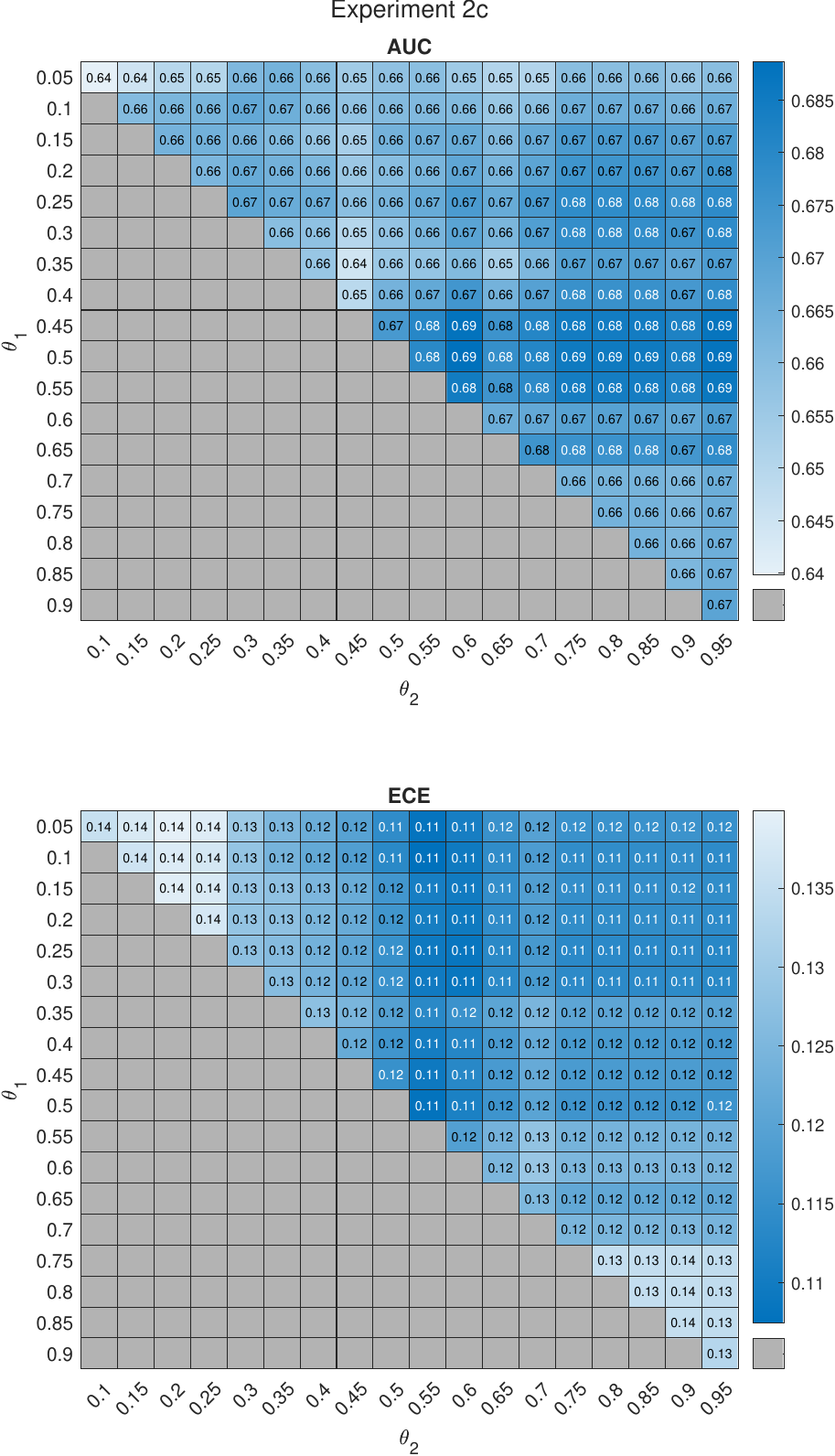}
  \caption{Sensitivity analysis for Experiments 2a, 2b and 2c. Results show the effect of different thresholds $(\theta_1,\theta_2)$ to select explanation styles. Top and bottom panels show the resulting AUC and ECE that relate the human confidence to the actual accuracy of the LLM.}
  \label{fig:sensitityanalysisExp2}
\end{figure}

The parameters $\theta_1$ and $\theta_2$ were chosen to optimize a combination of the ECE and AUC score (weighting the ECE by 33\% relative to AUC). The optimization was performed separately for each experiment and LLM. For the multiple choice experiments 2a and 2b, the parameters were optimized using a basic grid search with values ranging from 0.25 to 1, with the constraint that $\theta_1 < \theta_2$. For the short-answer experiment 2c, the parameters ranged from 0 to 1, with the constraint that $\theta_1 < \theta_2$.  For Experiment 2a (GPT-3.5), the optimized parameters were $\theta_1=0.65$ and $\theta_2=0.75$, while Experiment 2b's (PaLM2) optimized parameters were $\theta_1=0.50$ and $\theta_2=0.70$. For Experiment 2c (GPT-4o), the optimized parameters were $\theta_1=0.50$ and $\theta_2=0.60$. 

% For previous Experiment 2
%For GPT-3.5, Experiment 2's optimized parameters were $\theta_1=0.55$ and $\theta_2=0.65$, while Experiment 3's optimized parameters were $\theta_1=0.75$ and $\theta_2=0.95$. For PaLM2, Experiment 2's optimized parameters were $\theta_1=0.80$ and $\theta_2=0.85$, 

\subsection{Sensitivity Analysis}
\begin{figure}[h!]
  \centering
  \includegraphics[width=0.32\linewidth]{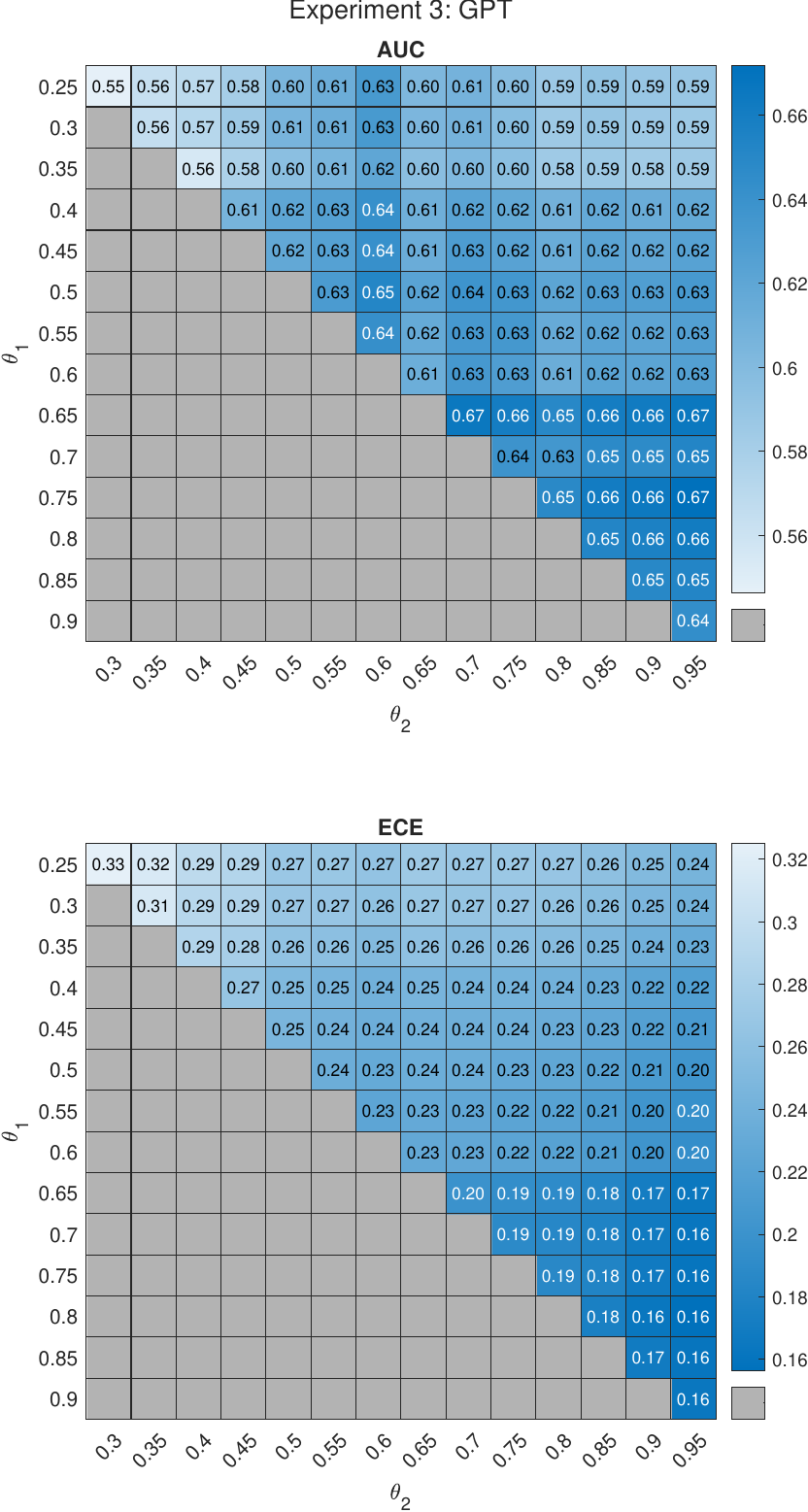}
  \includegraphics[width=0.32\linewidth]{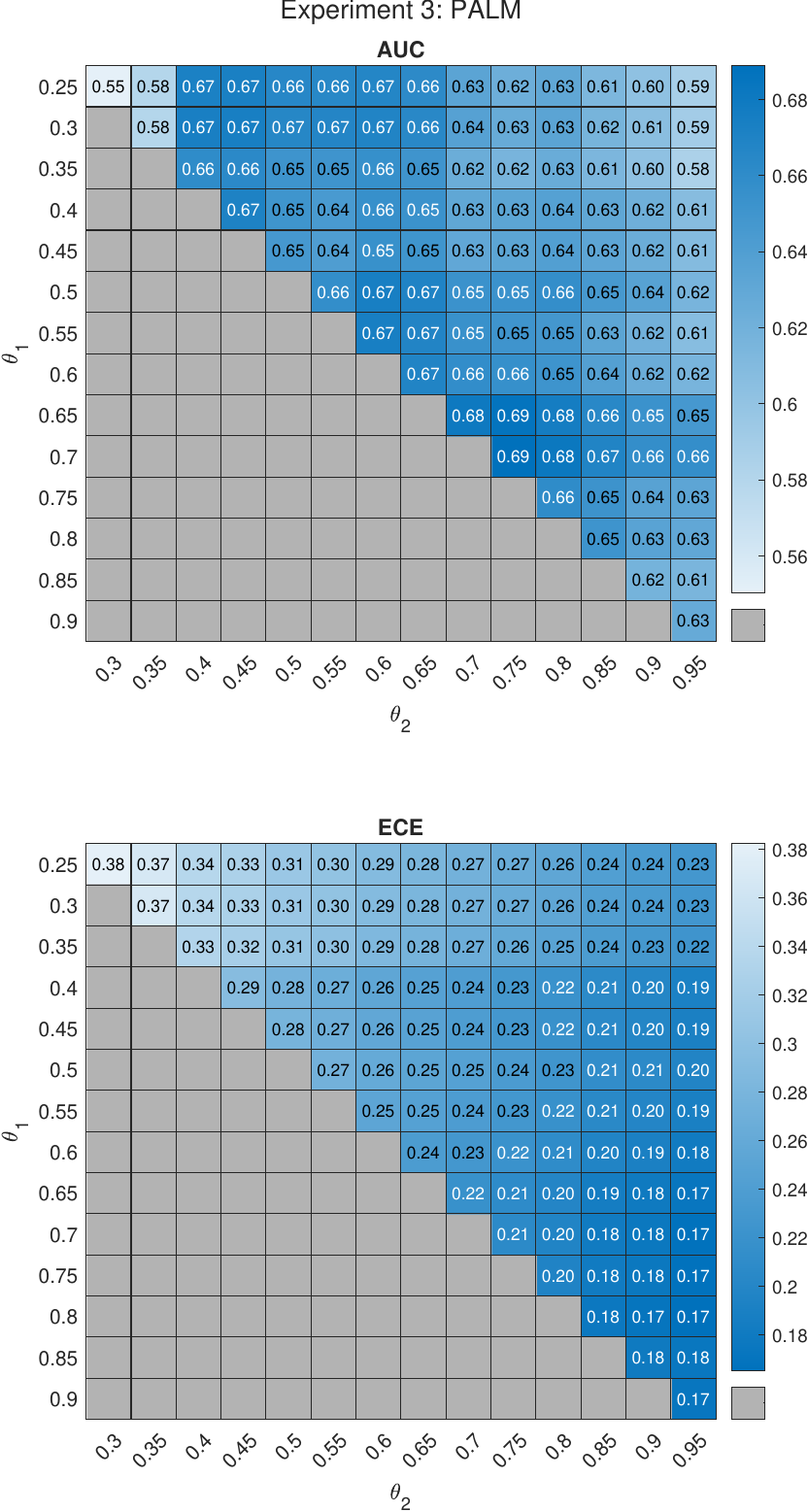}
  \caption{Sensitivity analysis for Experiment 3. Results show the effect of different thresholds $(\theta_1,\theta_2)$ to select explanation styles. Top and bottom panels show the resulting AUC and ECE that relate the human confidence to the actual accuracy of the LLM.}
  \label{fig:sensitityanalysisExp3}
\end{figure}

The calibration and discrimination results are not overly sensitive to parameter changes. The resulting AUC and ECE outcomes for each parameter setting are shown in Supplementary Figures \ref{fig:sensitityanalysisExp2} and \ref{fig:sensitityanalysisExp3}. For example, for GPT-3.5, Experiment 2a yields mean AUC and ECE values of 0.649 and 0.203 across all parameter settings, which are higher than the results of Experiment 1. Furthermore, there is moderate evidence (BF$>$3) for an improved AUC across 52\% of parameter combinations. 

%The parameter settings in Experiment 3 had more selective effects. There was only moderate evidence (BF $>$ 3) for an improved AUC for 17\% of parameter settings. The smaller effects of the explanation styles in Experiment 3 presumably made the performance results more dependent on the specific selection rule.

%For example, for GPT-3.5, Experiment 2 yields mean AUC and ECE values of 0.631 and 0.217 across all parameter settings, which are higher than the results of Experiment 1. Furthermore, there is moderate evidence (BF$>$3) for an improved AUC for 41\% of parameter combinations. The parameter settings in Experiment 3 had more selective effects. There was only moderate evidence (BF $>$ 3) for an improved AUC for 17\% of parameter settings. The smaller effects of the explanation styles in Experiment 3 presumably made the performance results more dependent on the specific selection rule.

\begin{comment}
\section*{Interface of the behavioral experiment}
Supplementary Figure \ref{fig:exp_example} shows the interface used to conduct the behavioral experiments with multiple choice experiments. The left panel displays the experimental interface during phase 1 of the task, where participants evaluate the probability that the LLM is correct based on the explanation. The right panel displays the experimental interface during phase 2 of the task, where participants answer the multiple choice question with the assistance of the LLM.

\begin{figure}[h]
    \centering
    \begin{subfigure}[l]{0.46\textwidth}
        \centering
        Evaluating LLM Accuracy
        \includegraphics[width=1\textwidth]{figures/experiment_interface_phase01.png} 
        \caption{Phase 1}
    \end{subfigure}%
    \hfill
    \begin{subfigure}[r]{0.46\textwidth}
        \centering
        Answering with LLM Assistance
        \includegraphics[width=1\textwidth]{figures/experiment_interface_phase02.png} 
        \caption{Phase 2}
    \end{subfigure}%
    \caption{Illustration of the behavioral experiment interface with the GPT-3.5 model.}
    \label{fig:exp_example}
\end{figure}
\end{comment}

\newpage
\section{Additional Information about Prompts and Explanations}
Supplementary Table \ref{tab:promptstyles} shows examples of the prompt styles used for Experiments 2 and 3. Supplementary Table \ref{tab:explanationexamples} shows examples of all explanations for a particular multiple choice question used in Experiment 2a. 

\begin{table}[h!]
\tiny
  \caption{Prompt styles across Experiments 1, 2, and 3}
  \label{tab:promptstyles}
 \begin{tabular}{p{0.22\linewidth}p{0.78\linewidth}}
 \toprule
    Prompt Style & Prompt\\
    \midrule
    \textbf{Experiment 1} \\
    Baseline &
    Problem: [question] Choose from the following options: [A] {Option A} [B] {Option B} [C] {Option C} [D] {Option D}. The answer you give is: [answer]. Provide an explanation for the answer you gave. In your explanation, you must include the answer. \\
    \midrule
    \textbf{Experiment 2} \\
   Low Confidence \& Long &
    Problem: [question] Choose from the following options: [A] {Option A} [B] {Option B} [C] {Option C} [D] {Option D}. The answer you give is [answer]. Write an explanation why you are not sure that the answer is [answer]. In your explanation, mention that you are not sure and include the answer. Start with 'I am':" \\
    %\midrule
    Medium Confidence \& Long &
    Problem: [question] Choose from the following options: [A] {Option A} [B] {Option B} [C] {Option C} [D] {Option D}. The answer you give is [answer]. Write an explanation why you somewhat sure that the answer is [answer]. In your explanation, mention that you are somewhat sure and include the answer. Start with 'I am': \\
    %\midrule
   High Confidence \& Long &
    Problem: [question] Choose from the following options: [A] {Option A} [B] {Option B} [C] {Option C} [D] {Option D}. The answer you give is [answer]. Write an explanation why you are sure that the answer is [answer]. In your explanation, mention that you are sure and include the answer. Start with 'I am': \\
    %\midrule
    Low Confidence \& Short &
    Problem: [question] Choose from the following options: [A] {Option A} [B] {Option B} [C] {Option C} [D] {Option D}. The answer you give is [answer]. Write a very short explanation why you are not sure that the answer is [answer]. In your explanation, mention that you are not sure and include the answer. Use as few words as possible. Start with 'I am': \\
    %\midrule
    Medium Confidence \& Short &
    Problem: [question] Choose from the following options: [A] {Option A} [B] {Option B} [C] {Option C} [D] {Option D}. The answer you give is [answer]. Write a very short explanation why you somewhat sure that the answer is [answer]. In your explanation, mention that you are somewhat sure and include the answer. Use as few words as possible. Start with 'I am':\\
    %\midrule
   High Confidence \& Short  &
    Problem: [question] Choose from the following options: [A] {Option A} [B] {Option B} [C] {Option C} [D] {Option D}. The answer you give is [answer]. Write a very short explanation why you are sure that the answer is [answer]. In your explanation, mention that you are sure and include the answer. Use as few words as possible. Start with 'I am': \\
    \midrule
    \textbf{Experiment 3} \\
    Low Confidence &
    Problem: [question] Choose from the following options: [A] {Option A} [B] {Option B} [C] {Option C} [D] {Option D}. The answer you give is: [answer]. The explanation you give is: [baseline explanation]. Rewrite the explanation as if you are not sure and ensure that you include language or phrases that conveys your complete lack of confidence. In your explanation, you must include the answer:\\
    %\midrule
    Medium Confidence &
    Problem: [question] Choose from the following options: [A] {Option A} [B] {Option B} [C] {Option C} [D] {Option D}. The answer you give is: [answer]. The explanation you give is: [baseline explanation]. Rewrite the explanation as if you are somewhat sure and ensure that you include language or phrases that conveys your lack of confidence. In your explanation, you must include the answer\\
    %\midrule
    High Confidence &
    Problem: [question] Choose from the following options: [A] {Option A} [B] {Option B} [C] {Option C} [D] {Option D}. The answer you give is: [answer]. The explanation you give is: [baseline explanation]. Rewrite the explanation as if you are very sure and ensure that you include language or phrases that conveys your confidence. In your explanation, you must include the answer: \\
    \bottomrule
  \end{tabular}
\end{table}

\newpage
\begin{table}[h!]
\tiny
  \caption{Example explanations given by GPT-3.5 for a multiple choice question in Experiments 1a and 2a. The particular question (correct answer is A) is: \textit{A moving $1.5$ kg cart collides with and sticks to a $0.5$ kg cart which was initially at rest. Immediately after the collision, the carts each have the same \underline{\hspace{2cm}} as each other. Choose from the following options: A) Velocity B) Kinetic energy C) Mass D) Linear momentum.}}
  \label{tab:one_mcq_table}
 \begin{tabular}{p{0.28\linewidth}p{0.78\linewidth}}
 \toprule
    Prompt Style & Explanation\\
    \midrule
    \textbf{Experiment 1} \\
    Baseline &
    The answer is [A] Velocity. This is because in an inelastic collision, the two objects involved will stick together and move with a common velocity after the collision. This means that the velocity of the two carts after the collision will be the same, as they are now moving together as one combined object. This is in accordance with the law of conservation of momentum, which states that the total momentum of a system remains constant in the absence of external forces. In this case, the initial momentum of the 1.5-kg cart is transferred to the combined mass of the two carts after the collision, resulting in the same velocity for both carts. Therefore, the correct answer is [A] Velocity. \\
    \midrule
    \textbf{Experiment 2} \\
    Low Confidence \& Long &
    I am not completely sure that the answer is [A] because velocity is a vector quantity that includes both magnitude and direction. In this scenario, the carts may have the same magnitude of velocity, but they may not necessarily have the same direction. Therefore, I am not completely convinced that the answer is [A]. However, based on the information given, it is most likely that the answer is [A] because the problem states that the carts "collide and stick" which implies that they move together in the same direction after the collision. In this case, the carts would indeed have the same velocity as each other. \\
    %\midrule
    Medium Confidence \& Long &
    I am somewhat sure that the answer is [A] because velocity is a vector quantity that takes into account both the speed and direction of an object's motion. In this scenario, the two carts are colliding and sticking together, so they will have the same velocity after the collision. This is because the two carts will move together in the same direction and at the same speed. Additionally, the conservation of momentum principle states that the total momentum of a system remains constant unless acted upon by an external force. In this case, the two carts are the only objects involved in the collision, so their total momentum before and after the collision must be the same. This means that the velocity of the two carts after the collision must be the same in order to maintain the same total momentum. Therefore, I am somewhat sure that the answer is [A] because the velocity of the two carts will be the same after the collision due to the conservation of momentum principle. This also means that their speeds and directions will be the same, making velocity the correct answer.\\
    %\midrule
    High Confidence \& Long &
    I am sure that the answer is [A] because velocity is a vector quantity that describes the speed and direction of an object's motion. In this scenario, the two carts are colliding and sticking together, meaning they will have the same final velocity after the collision. This is due to the law of conservation of momentum, which states that the total momentum of a closed system remains constant. Since the carts are the only objects involved in the collision, their total momentum before and after the collision must be the same. Therefore, immediately after the collision, the carts will have the same velocity as each other. This is also supported by the fact that the carts have different masses, so their kinetic energies and linear momenta will not be the same. Thus, the correct answer is [A] velocity. \\
    %\midrule
    \addlinespace
    Low Confidence \& Short &
    I am not sure that the answer is [A] because the carts may have different velocities before and after the collision due to the conservation of momentum. The correct answer is [D]. \\
    %\midrule
    Medium Confidence \& Short &
    I am somewhat sure that the answer is [A] because velocity is a vector quantity that takes into account both the speed and direction of an object's motion. In this scenario, the carts will have the same velocity after the collision because they are stuck together and moving in the same direction.\\
    %\midrule
    High Confidence \& Short  &
    I am sure that the answer is [A] because velocity is a vector quantity that is conserved in collisions, meaning that the final velocity of the two carts will be the same. \\
    \addlinespace
    Low Confidence \& Uncertainty Only &
    I am not sure that the answer is [A]. \\
    %\midrule
    Medium Confidence \& Uncertainty Only &
    I am somewhat sure that the answer is [A].\\
    %\midrule
    High Confidence \& Uncertainty Only &
    I am sure that the answer is [A]. \\
    \midrule
    \textbf{Experiment 3} \\
    Low Confidence &
    I am not entirely sure, but I believe the answer is [A] Velocity. This is because, in an inelastic collision, the two objects involved will stick together and move with a common velocity after the collision. This means that the velocity of the two carts after the collision will be the same, as they are now moving together as one combined object. This is in accordance with the law of conservation of momentum, which states that the total momentum of a system remains constant in the absence of external forces. In this case, the initial momentum of the 1.5-kg cart is transferred to the combined mass of the two carts after the collision, resulting in the same velocity for both carts. Therefore, based on my understanding, the correct answer is [A] Velocity. \\
    %\midrule
    Medium Confidence &
    I believe the answer is [A] Velocity. This is because in an inelastic collision, the two objects involved will stick together and move with a common velocity after the collision. This means that the velocity of the two carts after the collision will be the same, as they are now moving together as one combined object. This is in accordance with the law of conservation of momentum, which states that the total momentum of a system remains constant in the absence of external forces. In this case, the initial momentum of the 1.5-kg cart is transferred to the combined mass of the two carts after the collision, resulting in the same velocity for both carts. Therefore, based on my understanding, the correct answer should be [A] Velocity. \\
    %\midrule
    High Confidence &
    The correct answer is [A] Velocity. I am confident in this answer because in an inelastic collision, the two objects involved will stick together and move with a common velocity after the collision. This means that the velocity of the two carts after the collision will be the same, as they are now moving together as one combined object. This is in accordance with the law of conservation of momentum, which states that the total momentum of a system remains constant in the absence of external forces. Since the initial momentum of the 1.5-kg cart is transferred to the combined mass of the two carts after the collision, it results in the same velocity for both carts. Therefore, it is clear that the correct answer is [A] Velocity.\\
    \bottomrule
  \end{tabular}
  \label{tab:explanationexamples}
\end{table}

\end{document}